\documentclass[nohyperref]{article}

\usepackage{microtype}
\usepackage{graphicx}
\usepackage{subfigure}
\usepackage{booktabs} 

\usepackage{hyperref}

\usepackage[accepted]{icml2022}

\usepackage{amsmath}
\usepackage{amssymb}
\usepackage{mathtools}
\usepackage{amsthm}
\usepackage{amsfonts}
\usepackage{bm}

\usepackage[capitalize,noabbrev]{cleveref}

\theoremstyle{plain}
\newtheorem{theorem}{Theorem}[section]

\newtheorem{lemma}[theorem]{Lemma}

\theoremstyle{definition}
\newtheorem{definition}[theorem]{Definition}

\theoremstyle{remark}

\usepackage[textsize=tiny]{todonotes}

\usepackage{amsfonts}
\usepackage{bm}
\usepackage{xspace}
\usepackage{floatflt}

\usepackage{enumitem}
\setlist[enumerate]{
    topsep=0pt,itemsep=-1ex,partopsep=1ex,parsep=1ex,leftmargin=*
}
\setlist[itemize]{
    topsep=0pt,itemsep=-1ex,partopsep=1ex,parsep=1ex,leftmargin=*
}

\usepackage{thm-restate}

\usepackage{multirow}
\usepackage{makecell}

\icmltitlerunning{
    On the Equivalence Between Temporal and Static Equivariant Graph
    Representations
}

\begin{document}

\twocolumn[

\icmltitle{
    On the Equivalence Between \\
    Temporal and Static Equivariant Graph Representations
}

\icmlsetsymbol{equal}{*}

\begin{icmlauthorlist}
\icmlauthor{Jianfei Gao}{purdue}
\icmlauthor{Bruno Ribeiro}{purdue}
\end{icmlauthorlist}

\icmlaffiliation{purdue}{
    Department of Computer Science, Purdue University, West Lafayette, IN
    47906, USA
}

\icmlcorrespondingauthor{Jianfei Gao}{gao462@purdue.edu}
\icmlcorrespondingauthor{Bruno Ribeiro}{ribeiro@cs.purdue.edu}

\icmlkeywords{Temporal Graph Neural Network; Expressivity}

\vskip 0.3in
]



\printAffiliationsAndNotice{}

%

\newcommand{\graphthentime}{{\em graph-then-time}\xspace}
\newcommand{\timeandgraph}{{\em time-and-graph}\xspace}
\newcommand{\timethengraph}{{\em time-then-graph}\xspace}
\newcommand{\Graphthentime}{{\em Graph-then-time}\xspace}
\newcommand{\Timeandgraph}{{\em Time-and-graph}\xspace}
\newcommand{\Timethengraph}{{\em Time-then-graph}\xspace}
\newcommand{\onewlthentime}{{\em 1WLGNN-then-time}\xspace}

\newcommand{\timeandonewl}{{\em time-and-1WLGNN}\xspace}
\newcommand{\timethenonewl}{{\em time-then-1WLGNN}\xspace}
\newcommand{\Timeandonewl}{{\em Time-and-1WLGNN}\xspace}
\newcommand{\Timethenonewl}{{\em Time-then-1WLGNN}\xspace}

\newcommand{\timeandkwl}{{\em time-and-$k$WLGNN}\xspace}
\newcommand{\timethenkwl}{{\em time-then-$k$WLGNN}\xspace}
\newcommand{\Timeandkwl}{{\em Time-and-$k$WLGNN}\xspace}
\newcommand{\Timethenkwl}{{\em Time-then-$k$WLGNN}\xspace}

\newcommand{\timeandplus}{{\em time-and-GNN$^{+}$}\xspace}
\newcommand{\timethenplus}{{\em time-then-GNN$^{+}$}\xspace}
\newcommand{\Timeandplus}{{\em Time-and-GNN$^{+}$}\xspace}
\newcommand{\Timethenplus}{{\em Time-then-GNN$^{+}$}\xspace}

\newcommand{\exprlt}{\precneqq}
\newcommand{\exprgt}{\succneqq}
\newcommand{\exprleq}{\preceq}
\newcommand{\exprgeq}{\succeq}
\newcommand{\expreq}{\stackrel{e}{=}}

\newcommand{\tnX}{\boldsymbol{\mathsf{X}}}
\newcommand{\tnA}{\boldsymbol{\mathsf{A}}}
\newcommand{\stV}{\mathcal{V}}
\newcommand{\stE}{\mathcal{E}}
\newcommand{\fmR}{\mathbb{R}}
\newcommand{\fmG}{\mathbb{G}}
\newcommand{\fmT}{\mathbb{T}}
\newcommand{\tnS}{\boldsymbol{\mathsf{S}}}

\newcommand{\tnM}{\boldsymbol{\mathsf{M}}}
\newcommand{\tnZ}{\boldsymbol{\mathsf{Z}}}
\newcommand{\stN}{\mathcal{N}}
\newcommand{\tnH}{\boldsymbol{\mathsf{H}}}

\newcommand{\stU}{\mathcal{U}}
\newcommand{\stI}{\mathcal{I}}
\newcommand{\vcu}{\boldsymbol{u}}
\newcommand{\stF}{\mathcal{F}}

\newcommand{\vcx}{\boldsymbol{x}}
\newcommand{\vca}{\boldsymbol{a}}
\newcommand{\gpC}{\mathcal{C}}

\newcommand{\tnI}{\boldsymbol{\mathsf{I}}}
\newcommand{\tnJ}{\boldsymbol{\mathsf{J}}}
\newcommand{\tnE}{\boldsymbol{\mathsf{E}}}
\newcommand{\tnF}{\boldsymbol{\mathsf{F}}}
\newcommand{\bld}[1]{\tilde{#1}}

\begin{abstract}
This work formalizes the associational task of predicting node attribute
evolution in temporal graphs from the perspective of learning equivariant
representations.
We show that node representations in temporal graphs can be cast into two
distinct frameworks:
(a) The most popular approach, which we denote as \timeandgraph, where
equivariant graph (e.g., GNN) and sequence (e.g., RNN) representations are
intertwined to represent the temporal evolution of node attributes in the
graph;
and (b) an approach that we denote as \timethengraph, where the sequences
describing the node and edge dynamics are represented first, then fed as node
and edge attributes into a static equivariant graph representation that comes
after.
Interestingly, we show that \timethengraph representations have an expressivity
advantage over \timeandgraph representations when both use component GNNs that
are not most-expressive (e.g., 1-Weisfeiler-Lehman GNNs).
Moreover, while our goal is not necessarily to obtain state-of-the-art results,
our experiments show that \timethengraph methods are capable of achieving
better performance and efficiency than state-of-the-art \timeandgraph methods
in some real-world tasks, thereby showcasing that the \timethengraph framework
is a worthy addition to the graph ML toolbox.
\end{abstract}

\section{Introduction}
\label{sec:introduction}

Graph representation methods, in particular Graph Neural Networks
(GNNs)~\citep{%
    gori2005a,scarselli2005graph,duvenaud2015convolutional,gilmer2017neural%
} are part of powerful toolkits used in many real world applications~\citep{%
    battaglia2016interaction,fout2017protein,ying2018graph,chen2019supervised%
}.
GNNs have gathered a lot of attention among graph representation methods for
its ability to generate expressive node representations that are invariant to
node ordering in the graph, a.k.a. equivariant graph representations.
The equivariance of GNNs finds applications in predicting node and
graph labels, and in simulating dynamical systems~\citep{%
    battaglia2016interaction,xu2018how,morris2019weisfeiler,maron2019provably,%
    maron2019on,murphy2019relational,keriven2019universal,%
    dehmamy2019understanding%
}.

Despite the great power of GNNs to represent static graphs, many real world
graphs are temporal in nature, and static graph representations are thought to
be insufficient to embed such dynamics \citep{%
    berger-wolf2006a,fallani2014graph,ubaru2020dynamic%
}.
Hence, researchers have focused on combining static graph and sequence
representations together to deliver more powerful representations in order to
embed temporal graphs.
And this way, temporal graph neural networks (TGNNs) came to find applications
in domains as diverse as neuroscience~\citep{fallani2014graph,xu2019adaptive},
traffic forecasting~\citep{%
    yu2018spatio,cui2020traffic,zhao2020t,lv2020temporal%
}, disease spreading~\citep{deng2019graph,kapoor2020examining,gao2021stan},
social networks~\citep{berger-wolf2006a}, recommendation systems~\citep{%
    sankar2020dysat%
}, finance networks~\citep{pareja2020evolvegcn}, and crime analysis~\citep{%
    jin2020addressing%
}.

In most existing state-of-the-art TGNN works~\citep{%
    kapoor2020examining,gao2021stan,sankar2020dysat,cui2020traffic,zhao2020t,%
    lv2020temporal,pareja2020evolvegcn,jin2020addressing,manessi2020dynamic,%
    goyal2020dyngraph2vec%
}, the temporal graph is described as a sequence of graph snapshots:
These methods construct a graph representation for
each graph snapshot, and embed evolution of these node representations over
time as the final temporal graph representation.
We denote these \timeandgraph representations, which have dominated real world
applications in associational tasks.
Associational tasks are tasks that seek to predict node labels without
intervening on the system, in contrast to causal tasks which will consider
interventions.

Two recent works have proposed to treat temporal graph differently from the
\timeandgraph framework.
TGAT~\citep{xu2020inductive} memorizes all edges from the past snapshots and
constructs a static heterogeneous graph, then extracts representations from the
constructed graph as final TGNN representations.
TGN~\citep{rossi2020temporal} extends TGAT, and uses sequence representations
of all edges connecting to the same nodes to replace node attributes in the
heterogeneous graph.
Both works achieve state-of-the-art performance on social and finance networks,
providing an alternative method to represent temporal graphs other than
\timeandgraph.

In this paper, we generalize these recent efforts and propose an alternative
representation framework for temporal graphs: \Timethengraph, which first
sequentially represents the temporal evolution of node and edge attributes.
Then, we use these temporal representations to define a static graph
representation.
We theoretically prove that \timethengraph architectures that use GNNs with
expressivity limited to 1-WL power~\cite{xu2018how,morris2019weisfeiler} as
building blocks have an expressivity advantage over \timeandgraph architectures
that also use the same 1-WL GNN architecture as building blocks.
Our experiments show that \timethengraph can also hold an empirical advantage
over \timeandgraph in real-world applications.

{\bf Contributions.}
Our work introduces \timethengraph representations and studies expressive power
over temporal graph representations.
Our contributions are as follows:
\begin{enumerate}
\item
\Timethengraph representations are more expressive than \timeandgraph
representations if the sequence representation is most expressive (e.g.,
RNN, transformer) and the graph representation is a standard GNN~\citep{%
    kipf2016semi,velickovic2018graph%
}, or more precisely, a 1-Weisfeller-Lehman GNN as \citet{%
    xu2018how,morris2019weisfeiler,maron2019provably%
}.
\Timethengraph and \timeandgraph representations are equally expressive when
both the temporal sequence and graph representations are most expressive (e.g.,
the graph representations of \citet{maron2019on,murphy2019relational}).
\item
Our experiments confirm that our \timethengraph methods outperform
state-of-the-art \timeandgraph, TGAT and TGN methods in a specific synthetic
task, and obtain better or equivalent performance and efficency in all
real-world applications.
\end{enumerate}

\section{\Timeandgraph and \Timethengraph}
\label{sec:prelim}

{\bf Background.}
The definition of temporal graph is an extension of that of static graph, thus
we introduce static graphs first.

\begin{definition}[Static graph]
\label{def:stat_graph}
A static graph can be defined as $(\tnX, \tnA)$, where $
    \tnX \in \fmR^{|\stV| \times p}
$ are the node attributes and $
    \tnA \in \fmR^{|\stV| \times |\stV| \times p}
$ are the edge attributes, with $\stV$ the set of unique nodes in the graph and
$p \geq 1$ the dimension of the observed node and edge attributes.
\end{definition}

We denote the family of all static graphs satisfying the definition as $
    \fmG_{|\stV|, p} := \fmR^{|\stV| \times p} \times
    \fmR^{|\stV| \times |\stV| \times p}
$.

In this paper, we will use two different but equivalent forms to describe a
temporal graph:
The {\em snapshotted} form (\Cref{def:temp_graph_snap}) and {\em aggregated}
form (\Cref{def:temp_graph_aggr}).
We provide a sketched example to help understand two forms in \Cref{%
    fig:temp_graph%
}.

\begin{definition}[Snapshotted temporal graph]
\label{def:temp_graph_snap}
A temporal graph over $T$ times can be defined as a sequence of static
graph snapshots $
    \big[ \big( \tnX_{:, t}, \tnA_{:, :, t} \big) \big]_{t = 1}^{T}
$, where $\tnX \in \fmR^{|\stV| \times T \times p}$ are the node attributes and
$\tnA \in \fmR^{|\stV| \times |\stV| \times T \times p}$ are the edge
attributes over all $T$ times, with $\stV$ the set of all possible unique nodes
in the temporal graph and $p \geq 1$ is the dimension of the observed node and
edge attributes.
\end{definition}

\begin{definition}[Aggregated temporal graph]
\label{def:temp_graph_aggr}
A temporal graph over $T$ time steps can alternatively be defined as a static
graph $\big( \tnX_{:, \leq T}, \tnA_{:, :, \leq T} \big)$, where $
    \tnX \in \fmR^{|\stV| \times T \times p}
$ are the node attributes and $
    \tnA \in \fmR^{|\stV| \times |\stV| \times T \times p}
$ are the edge attributes over all $T$ times, with $\stV$ the set of all
possible unique nodes in the temporal graph.
Here, $\tnX_{:, \leq T} = \big[ \tnX_{:, 1}, \cdots, \tnX_{:, T} \big]$, $
    \tnA_{:, :, \leq T} = \big[ \tnA_{:, :, 1}, \cdots, \tnA_{:, :, T} \big]
$ integrate all past node and edge attributes as sequences of standalone
attributes, and $p \geq 1$ is the dimension of the observed node and edge
attributes (attributes may be filled by null if not present).
\end{definition}

\begin{figure}[tb]
\begin{center}
\centerline{\includegraphics[width=\columnwidth]{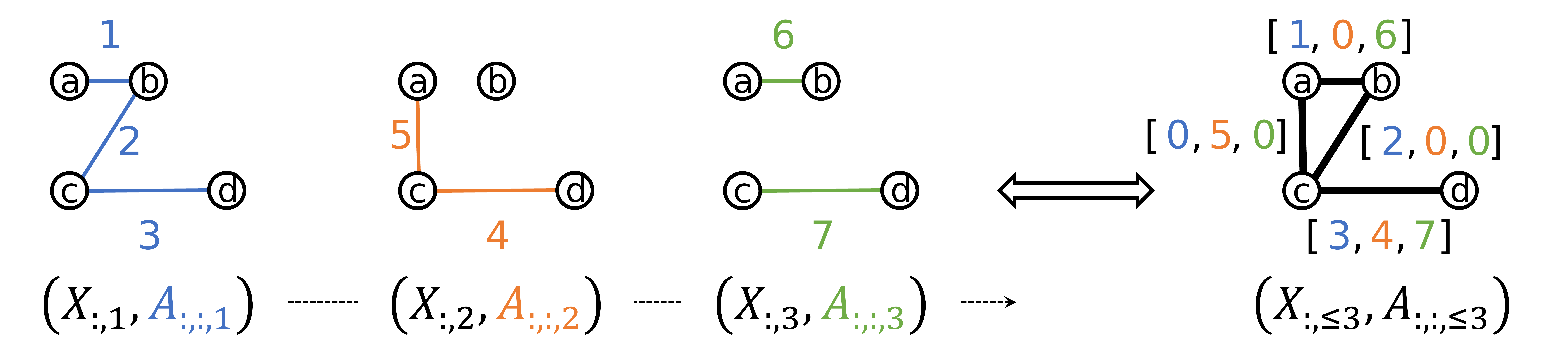}}
\caption{
    Equivalent snapshotted (left) and aggregated (right) form examples of a
    3-time temporal graph without node attributes.
    Aggregated form collects all edges appeared at least once in snapshotted
    form, and concatentate corresponding edge attributes into sequences as
    aggregated edge attributes.
    If an edge attribute is missing in the concatenation, it will be
    padded by null attribute 0.
}
\label{fig:temp_graph}
\end{center}
\vskip -0.3in
\end{figure}

We denote the space of all temporal graphs satisfying either definitions as $
    \fmT_{|\stV|, T, p} := \fmR^{|\stV| \times T \times p} \times
    \fmR^{|\stV| \times |\stV| \times T \times p}
$.

Next, we will introduce two atomic representations used in both \timeandgraph
and \timethengraph:
GNNs and Recurrent Neural Networks (RNNs).

Our GNN representations follow the Message Passing Neural Network~\citep{%
    gilmer2017neural%
} schema on arbitrary static graph $(\tnX, \tnA) \in \fmG_{|\stV|, p}$.
Formally, the $l$-th layer of a GNN is defined as:
\begin{equation}
\label{eqn:gnn}
\begin{aligned}
\tnM_{i}^{(l)} & = \sum\limits_{j \in \stN(i)} \text{MSG}^{(l)}\big(
    \tnZ_{i}^{(l - 1)}, \tnZ_{j}^{(l - 1)}, \tnA_{i, j}
\big),
\\
\tnZ_{i}^{(l)} & = \text{UPDATE}^{(l)}\big(
    \tnZ_{i}^{(l - 1)}, \tnM_{i}^{(l)}
\big),
\end{aligned}
\end{equation}
where $\tnZ_{i}^{(l)}$ represents the embedding of arbitrary node $i \in \stV$
obtained at layer $l$ and $\stN(i)$ defines the neighbor set of $i$.
We also initialize the corner case $\tnZ_{i}^{(0)} = \tnX_{i}$.
$\text{MSG}^{(l)}$ and $\text{UPDATE}^{(l)}$ are arbitrary learnable functions,
e.g., Multi-Layer Perceptions (MLPs).
For the ease of notation, we denote a stack of $L$ GNN layers by
\begin{align}
\tnZ^{(L)} = \text{GNN}^{L}(\tnX, \tnA), \label{eqn:gnn_stack}
\end{align}
where all the variables are as in \Cref{eqn:gnn}.
This definition covers widespread GNNs such as GCN~\citep{kipf2016semi},
GAT~\citep{velickovic2018graph}, and GIN~\citep{xu2018how}.

Besides GNN as graph representations, sequence representations are also
essential in \timeandgraph and \timethengraph.
In this work we will generally refer to all sequence representation methods as
Recurrent Neural Network (RNN) schema which covers widespread methods, e.g.,
GRU~\citep{cho2014learning,chung2014empirical} or LSTM~\citep{%
    hochreiter1997long%
}.
Strictly speaking, RNN recursively embeds the current and past observations for
arbitrary sequence $\tnS \in \fmR^{T \times p}$ where $p \geq 1$ defines
dimension of elements, and is formally defined as:
\begin{align}
\tnH_{t} = \text{Cell}\big( \tnS_{t}, \tnH_{t - 1} \big) , \label{eqn:rnn}
\end{align}
where $\tnH_{t}$ represents the embedding of sub-sequence $\tnS_{1:t}$ for $
    1 \leq t \leq T
$.
We initialize the corner case $\tnH_{0} = 0$.
$\text{Cell}$ is arbitrary learnable function, e.g., GRU cell.
For the ease of notation, we denote RNN on the full sequence as:
\begin{align}
\tnH_{T} = \text{RNN}(\tnS).
\end{align}
We also consider transformer and other self-attention mechanism
methods~\citep{vaswani2017attention,bahdanau2014neural}, where we define $
    \tnH_{T}
$ as obtained by a set representation instead.

Finally, we will define \timeandgraph and \timethengraph methods based on
aforementioned input domain and atomic representation concepts.

{\bf \Timeandgraph.}
\Timeandgraph is the most common representation adopted in the literature.
The most widely used \timeandgraph representations (e.g., \citet{%
    li2019predicting,chen2018gc,seo2018structured%
}) will embed snapshotted temporal graph $
    \big[ \big( \tnX_{:, t}, \tnA_{:, :, t} \big) \big]_{t = 1}^{T}
$.
First, 1-WL GNNs are independently applied to each snapshot $
    \big( \tnX_{:, t}, \tnA_{:, :, t} \big)
$.
Then, the outputs of these GNNs are concatentated and embedded into sequence
representations, giving the final representation of all nodes for the temporal
graph.
\Timeandgraph is formally abstracted as:
\begin{equation}
\label{eqn:timeandgraph}
\begin{aligned}
\tnH_{i, t} = \text{Cell}\bigg(
    & \Big[
        \text{GNN}_\text{in}^{L}\big(
            \tnX_{:, t}, \tnA_{:, :, t}
        \big)
    \Big]_{i},
    \\
    & \Big[
        \text{GNN}_\text{rc}^{L}\big(
            \tnH_{:, t - 1}, \tnA_{:, :, t}
        \big)
    \Big]_{i}
\bigg),
\end{aligned}
\end{equation}
where $\tnH_{i, t}$ is \timeandgraph representation of temporal node $
    i \in \stV
$ at time $1 \leq t \leq T$.
We initialize the corner case $\tnH_{i, 0} = 0$ for any node $i$.
$\text{GNN}_\text{in}^{L}$ encodes each snapshot inputs, $
    \text{GNN}_\text{rc}^{L}
$ encodes represetations from historical snapshots, and $\text{Cell}$ is a RNN
cell embedding evolutions of those GNN representations.
For an arbitrary temporal graph, the outputs at the last step $\tnH_{i, T}$ is
regarded as the final temporal graph representation of node $i \in \stV$.

There are several well-known temporal graph works, e.g., \citet{%
    manessi2020dynamic,sankar2020dysat,seo2018structured%
}, that fall into a strict subset of defined \timeandgraph methods where $
    \text{GNN}_\text{rc}^{L}
$ directly outputs its node input $\tnH_{:, t - 1}$.
We specially call this subset \graphthentime, and it is defined as:
\begin{align}
\tnH_{i, t} = \text{Cell}\bigg(
    \Big[
        \text{GNN}_\text{in}^{L}\big(
            \tnX_{:, t}, \tnA_{:, :, t}
        \big)
    \Big]_{i},
    \tnH_{i, t - 1}
\bigg)
\label{eqn:graphthentime}
\end{align}

{\bf \Timethengraph.}
This work proposes the \timethengraph framework, which is a more powerful
representation than \timeandgraph for temporal graphs using 1-WL GNNs.
Instead of applying GNNs over each snapshot, we represent the evolution of node
attributes using a sequence model.
We also perform another sequence representation for the temporal edge attribute
evolution.
These new node and edge representations become node and edge attributes in
a new (static) graph, which is then encoded by a GNN.
Thus, our representation architecture will embed aggregated temporal graph $
    \big( \tnX_{:, \leq T}, \tnA_{:, :, \leq T} \big)
$, and it is formally defined as:
\begin{equation}
\label{eqn:timethengraph}
\begin{aligned}
\tnH^\text{node}_{i} & = \text{RNN}^\text{node}\big( \tnX_{i, \leq T} \big),
\quad \forall i \in \stV,
\\
\tnH^\text{edge}_{i, j} & = \text{RNN}^\text{edge}\big(
    \tnA_{i, j, \leq T}
\big), \quad \forall (i, j) \in \stE,
\\
\tnZ & = \text{GNN}^{L}\big( \tnH^\text{node}, \tnH^\text{edge} \big).
\end{aligned}
\end{equation}
where $\tnH^\text{node}$ and $\tnH^\text{edge}$ are the representations of node
and edge evolutions, $\stE$ is the set of all possible edges in aggregated form
and $\tnZ$ is final temporal graph representation of all nodes.

State-of-the-art temporal graph representations TGAT and TGN can be treated as
special cases of \timethengraph, where $\text{RNN}^\text{node}$ and $
    \text{RNN}^\text{edge}
$ only takes observable nodes and edges as inputs (subsets of $
    \tnX_{:, \leq T}
$ and $\tnA_{:, :, \leq T}$ with only non-null attributes).

When contrast against \timeandgraph, our proposed \timethengraph framework
provides following advantages:

{
    \bf 1. \Timethengraph with 1-WL GNNs is more expressive than \timeandgraph
    with 1-WL GNNs.
}
\Timethengraph can distinguish different temporal graphs which will always be
the same on \timeandgraph representation space with 1-WL GNNs.
Besides, we show that for any \timeandgraph representation, there will be an
equivalent counterpart in \timethengraph representation.
The expressivity advantage of distinguishing more temporal graphs is
empirically beneficial to real-world applications utilizing these
representations.
We will introduce more details about expressvity in next section.

{\bf 2. \Timethengraph can be more computationally efficient in practice.}
GNNs often operate on sparse temporal graphs, and could benefit a lot from
parallel computation.
In \timeandgraph framework, there is a bottleneck that $
    \text{GNN}^{L}_\text{rc}
$ must wait for $\tnH_{:, t - 1}$ until all previous snapshots are processed.
Thus, \timeandgraph must process snapshots one-by-one, which does not
fully utilize parallelization.
In contrast, there is not such bottleneck in \timethengraph, both RNNs and GNNs
can compute many of the embeddings in parallel during their {\em time} and {
        \em graph
} phase, respectively.
We empirically show that our \timethengraph methods train faster than all
baselines in real-world tasks when the number of temporal edges is not large.

\section{Expressivity of Equivariant Temporal Graph Representations}
\label{sec:express}

In this section, we will discuss more details about the expressivity advantage
of \timethengraph over \timeandgraph.
First, however, we will clarify the concept of expressivity.

In the following analysis, we will treat the representation of an arbitrary
domain $\stU$ as a function $f: \stU \rightarrow \fmR^{d}$ which embeds any
element from $\stU$ into a low-dimensional vector on $\fmR^{d}$, where $
    d \geq 1
$ is the representation space dimensionality.
For example, a specific GNN architecture defines a family of representations $
    \stF
$ for domain $\stU = \fmG_{|\stV|, p}$, and a specific function $f \in \stF$
defines a specific set of parameters for such GNN architecture.
Similarly, TGNNs defines a family of representations $\stF'$ representations
for domain $\stU = \fmT_{|\stV|, T, p}$.

In static and temporal graph expressvity studies~\citep{%
    xu2018how,maron2019on,morris2019weisfeiler,murphy2019relational,%
    azizian2020characterizing,kileel2019on,kapoor2020examining%
}, the expressivity of a representation $f$ for an arbitrary domain $\stU$ is
represented by the cardinality of distinguishable inputs on its representation
space.
To help understand this concept, we describe {\em identifiable set}, which is
then used to define the expressivity of representations on arbitrary domain.

%
\begin{definition}[Identifiable Set]
\label{def:identify_set}
The identifiable set $\stI(f, \stU)$ of a representation $f$ with domain $\stU$
is a set s.t.:
\begin{enumerate}
\item
It is a subset of $\stU$ that is $\stI(f, \stU) \subseteq \stU$;
\item
None of its elements have the same representations, that is $
    \forall \vcu_{1} \neq \vcu_{2} \in \stI( f, \stU), f\big( \vcu_{1} \big)
    \neq f\big( \vcu_{2} \big)
$;
\item
It contains all unique representations over domain $\stU$, that is, $
    \forall \vcu_{1} \in \stU \backslash \stI(f, \stU), \exists \vcu_{2} \in
    \stI(f, \stU), f\big( \vcu_{1} \big) = f\big( \vcu_{2} \big)
$.
\end{enumerate}
\end{definition}
In other words, $\stI(f, \stU)$ is the maximal subset of $\stU$ that $f$ can
differentiate any pair of elements in that domain.
We may have multiple identifiable sets satisfying \Cref{def:identify_set} for
the same representation $f$ with domain $\stU$.
We find that for any pair of such identifiable sets, there are always bijection
between them.
Thus, all satisfying identifiable sets are equivalent in the sense of
cardinality on representation space, thus we can use an arbitrary identifiable
set for later expressivity analysis.
We provide a proof of the bijection in \Cref{subsec:proof_biject}.

{\bf Node isomorphism.}
When studying the second condition of \Cref{def:identify_set} in graph domains,
where $\vcu_1,\vcu_2$ are two graphs, we define equality $\vcu_1 = \vcu_2$ as
having graphs $\vcu_1$ and $\vcu_2$ being isomorphic.
We say two temporal graphs are isomorphic if their aggregation forms (\Cref{%
    def:temp_graph_aggr%
}) are isomorphic.

We now define the expressivity of representations on an arbitrary domain $\stU$
based on identifiable set.
Since we eventually want to compare the expressivity between \timeandgraph and
\timethengraph which are two families of representations, we formally quantify
the expressivity comparison between two arbitrary representation familes on the
same domain.

\begin{figure}[tb]
\begin{center}
\centerline{\includegraphics[width=\columnwidth]{fig_dyncsl}}
\caption{
    {
        \bf A synthetic task where only \timethengraph is expressive enough
        with 1-WL GNNs.
    }
    The top and bottom 2-time temporal graphs on the left side has snapshots of
    different structure at time $t_{2}$ (denote by $\gpC_{7, 2}$ and $
        \gpC_{7, 1}
    $).
    The two aggregated temporal graphs on the right side are corresponding
    aggregation of two snapshotted forms, where edge attributes are sequences
    of non-existing (0) and existing (1) over time, denoted by different
    colors.
    The task is to distinguish structure difference between top and bottom
    temporal graphs.
    \Timeandgraph using 1-WL GNNs will fail since GNN always output the same
    node representation for both snapshot $\gpC_{7, 2}$ and $\gpC_{7, 1}$ (see
    \citep{murphy2019relational}).
    On the other hand, \timethengraph works since the aggregated forms already
    show different aggregated topologies that any 1-WL GNN can easily
    distinguish.
}
\label{fig:dyncsl}
\end{center}
\vskip -0.3in
\end{figure}

\begin{definition}[Quantifying Levels of Expressivity]
\label{def:express_more}
Consider two representation families $\stF_{1}$ and $\stF_{2}$ of domain
$\stU$.
We say $\stF_{1}$ is more or as equally expressive as $\stF_{2}$, if and only
if $\exists f_{1} \in \stF_{1}$, $
    \forall f_{2} \in \stF_{2}, \big\vert \stI \big( f_{2}, \stU \big)
    \big\vert \leq \big\vert \stI \big( f_{1}, \stU \big) \big\vert
$.
We denote this by $\stF_{2} \exprleq_{\stU} \stF_{1}$.
\\
It is strictly more expressive, $\stF_{2} \exprlt_{\stU} \stF_{1}$, if and
only if $\exists f_{1} \in \stF_{1}$, $
    \forall f_{2} \in \stF_{2}, \big\vert \stI \big( f_{2}, \stU \big)
    \big\vert < \big\vert \stI \big( f_{1}, \stU \big) \big\vert
$.
\\
If two representation families satisfies both $
    \stF_{2} \exprleq_{\stU} \stF_{1}
$ and $\stF_{1} \exprleq_{\stU} \stF_{2}$, we say that they are equally
expressive, $\stF_{2} \expreq_{\stU} \stF_{1}$.
\end{definition}

\begin{definition}[Most Expressive]
\label{def:express_most}
A representation family $\stF^{+}$ for $\stU$ is the most expressive
representation family if and only if $\forall \vcu_{1}, \vcu_{2} \in \stU$, $
    \exists f^{+} \in \stF^{+}$, $f^{+}\big( \vcu_{1} \big) = f^{+}\big(
        \vcu_{2}
    \big) \Longleftrightarrow \vcu_{1} = \vcu_{2}
$.
We call such $f^{+}$ as the most expressive representation function.
\end{definition}

Generally speaking, more expressive representation family means that the
architecture $\stF$ is being able to better distinguish distinct elements in
domain $\stU$, and most expressive representation family means being able to
distinguish all distinct elements in domain $\stU$.

\begin{lemma}[Expressivity \& Representations]
\label{lem:express_more_alter}
For two representation families $\stF_{1}$ and $\stF_{2}$ of domain $\stU$, if
for any function of $\stF_{2}$, there is an equivalent function on $\stU$ in $
    \stF_{1}
$, then $\stF_{2} \exprleq_{\stU} \stF_{1}$, that is,
\begin{equation*}
\begin{aligned}
& \forall f_{2} \in \stF_{2}, \exists f_{1} \in \stF_{1}, \text{ s.t. }
\forall \vcu \in \stU, f_{1}(\vcu) = f_{2}(\vcu)
\\
& \Longrightarrow \stF_{2} \exprleq_{\stU} \stF_{1}.
\end{aligned}
\end{equation*}
\end{lemma}
\Cref{lem:express_more_alter} is straightforward: If $\stF_{1}$ can
simulate any representation function in $\stF_{2}$ on domain $\stU$, then it
distinguish at least the same elements.
It also reveals a more direct way to compare expressivity between
representation families than cardinality, and we will adopt this in the
expressivity comparison between \timeandgraph and \timethengraph.

Now that we have formally defined the concept of expressivity, and can
analyze the expressivity of \timeandgraph and \timethengraph frameworks.
To start, we need to understand the expressivity of their components, GNNs and
RNNs, which has already been sufficiently explored in the literature.
Thus, we can directly take expressivity conclusions from existing works:
\begin{enumerate}
\item
Common message passing GNNs including GCN~\citep{kipf2016semi},
GAT~\citep{velickovic2018graph}, and GIN~\citep{xu2018how} are the same
expressive as 1-Weisfeiler-Lehman (1-WL) test~\citep{%
    leman1968reduction,douglas2011the,xu2018how,morris2019weisfeiler%
}, which are NOT the most expressive graph representation.
\item
There are more expressive GNNs than 1-WL GNNs, e.g., k-WL GNNs~\citep{%
    morris2019weisfeiler}.
The most expressive GNNs, denoting as GNN$^{+}$, also exist~\citep{%
    maron2019on,murphy2019relational%
}.
However, all of them are not as efficient as 1-WL, thus is not widely used.
\item
RNNs can be most expressive sequence representations~\citep{
    siegelmann1995on,yun2020are%
}.
GRU~\citep{chung2014empirical}, LSTM~\citep{hochreiter1997long} and
transformer~\citep{vaswani2017attention} architectures are most expressive
RNNs in this paper.
\end{enumerate}

To differentiate temporal graph representations using GNNs of different
expressivities, we replace the graph by exact type of GNNs, e.g.,
\timethengraph with 1-WL GNNs is denoted as \timethenonewl.
We derive the expressivity of temporal graphs based on aforementioned
conclusions.

\begin{restatable}{theorem}{timeonewl}[Temporal 1-WL Expressivity]
\label{thm:temp_1wl_express}
\Timethengraph is strictly more expressive than \timeandgraph representation
family on $\fmT_{|\stV|, T, p}$ when the graph representation is a 1-WL GNN:
\begin{align*}
& \text{\onewlthentime} \exprlt_{\fmT_{|\stV|, T, p}} \text{\timeandonewl}
\\
& \exprlt_{\fmT_{|\stV|, T, p}} \text{\timethenonewl}.
\end{align*}
\end{restatable}

In \Cref{subsec:proof_temp_1wl_express}, we prove this by showing that we can
always construct a \timethengraph representation that outputs the same
embeddings as an arbitrary \timeandgraph representation.
Thus, by \Cref{lem:express_more_alter}, \timethengraph is as expressive as
\timeandgraph.
We also provide a task in \Cref{fig:dyncsl} where any \timeandgraph
representation will fail while a \timethengraph would work, which then, added
to the previous result, proves that \timethengraph is strictly more expressive
than \timeandgraph.

\begin{restatable}{theorem}{timegraph}[Temporal GNN Expressivity]
\label{thm:temp_graph_express}
Using more expressive GNNs, e.g., kWL GNNs, will improve the expressivity of
\timethengraph, and if a most expressive GNN$^+$ is used, \timethengraph and
\timeandgraph representation families will both be most expressive:
\begin{equation*}
\begin{aligned}
& \text{\timethenonewl} \exprlt_{\fmT_{|\stV|, T, p}} \text{\timethenkwl}
\\
& \exprlt_{\fmT_{|\stV|, T, p}} \text{\timethenplus}
\expreq_{\fmT_{|\stV|, T, p}} \text{\timeandplus}.
\end{aligned}
\end{equation*}
\end{restatable}

A proof of this is also provided in \Cref{subsec:proof_temp_graph_express}.

\Cref{thm:temp_1wl_express,thm:temp_graph_express} show that
\timethengraph is more expressive than \timeandgraph as long as we must use
1-WL GNNs, but they are indeed equivalent when most expressive GNNs are
available.

\section{Related Work}
\label{sec:related}

{\bf \Timeandgraph.}
GCRN-M2~\citep{seo2018structured} is the first work, as far as we know, to
adopt \Cref{eqn:timeandgraph} combining SpectralGCN~\citep{%
    defferrard2016convolutional%
} and LSTM.
DCRNN~\citep{li2018diffusion} is a \timeandgraph application in traffic
forecasting utilizing SpectralGCN and GRU.
LRGCN~\citep{li2019predicting} is another similar work but has further
processing on node representations.
GGNN~\citep{li2016gated} is a \timeandgraph application where node
attributes do not change, thus it removes $\text{GNN}_\text{in}$ that
most \timeandgraph methods have.

{\bf \Graphthentime.}
The framework defined by \Cref{eqn:graphthentime} is the most popular among
both traditional and state-of-the-art works for its simplicity and efficiency.
\citet{seo2018structured} proposes GCRN-M1 in the same paper of GCRN-M2 but
GCRN-M1 uses a \graphthentime framework.
\citet{manessi2020dynamic} also proposes something similar, WD-GCN, which
combines GCN and LSTM through a \graphthentime framework.
Another state-of-the-art work, DySAT~\citep{sankar2020dysat}, exploits
self-attention mechanism of GAT~\citep{velickovic2018graph} and a
transformer~\citep{vaswani2017attention} via graph and sequence components.
Of special interest is EvolveGCN~\citep{pareja2020evolvegcn}, which learns
the temporal dynamics through the GCN parameters rather than node
representations.
In EvolveGCN, GCN parameters are passed as recurrent states instead of
historical embeddings.
DynGEM~\citep{goyal2018dyngem} is similar to EvolveGCN, but it directly
inherits parameters from previous step instead of an RNN to model
parameter evolution.

{\bf \Timethengraph.}
To the best of our knowledge, the first \timethengraph work is \citet{%
    rahman2018dylink2vec%
}, which only uses non-parametric graph and sequence components.
Two other works, TGAT~\citep{xu2020inductive} and TGN~\citep{%
    rossi2020temporal%
}, are existing state-of-the-art \timethengraph works.
Instead of accumulating even non-existing edge attributes, they simply collect
all observed edges in past snapshots, and extract static graph representations
from a heterogeneous graph constructed through those edges.
THINE~\citep{huang2021temporal} and CAW~\citep{wang2021inductive} are works
close to TGAT but use random walks for label propagation instead of GNNs.
\citet{makarov2021temporal} uses the same procedure as CAW but applied to TGN.
Our \timethengraph method differ from the above methods, since we distill the
\timethengraph framework into its basic components:
A powerful temporal representation (e.g., a GRU) and a good 1-WL GNN graph
representation.

\begin{table}[t]
\caption{
    {\bf Statistics of datasets.}
    We collect and denote the number of samples (temporal graphs) as $N$,
    number of nodes as $|\stV|$, minimum and maximum number of edges for all
    snapshots as $\min_{t}\big| \stE_{t} \big|$ and $
        \max_{t}\big| \stE_{t} \big|
    $, number of snapshots per sample (temporal graph) as $T$, node and edge
    attribute dimensionality as $
        \big| p_{\tnX} \big|, \big| p_{\tnA} \big|
    $, learning target dimensionality as $|\mathbf{y}|$ and number of graphs
    per minibatch as $B$.
}
\label{tab:meta}
%
\begin{center}
\resizebox{\columnwidth}{!}{
\begin{tabular}{lrrrrrrrrr}
\hline
\multicolumn{1}{c}{Dataset} & \multicolumn{1}{r}{$N$} &
\multicolumn{1}{r}{$|\stV|$} &
\multicolumn{1}{r}{$\min_{t}\big| \stE_{t} \big|$} &
\multicolumn{1}{r}{$\max_{t}\big| \stE_{t} \big|$} & \multicolumn{1}{r}{$T$} &
$\big| p_{\tnX} \big|$ & $\big| p_{\tnA} \big|$ & $|\mathbf{y}|$ & $B$ \\
\hline
DynCSL & 200 & 19 & 76 & 76 & 8 & 0 & 0 & 1 & 1 \\
Brain10 & 1 & 5000 & 154094 & 167944 & 12 & 20 & 0 & 1 & 1 \\
\hline
PeMS04 & 16980 & 307 & 680 & 680 & 12 & 5 & 1 & 3 & 256 \\
PeMS08 & 17844 & 170 & 548 & 548 & 12 & 5 & 1 & 3 & 256 \\
Spain-COVID & 443 & 52 & 7030 & 7030 & 7 & 1 & 2 & 1 & 64 \\
English-COVID & 54 & 129 & 836 & 2158 & 7 & 1 & 1 & 1 & 4 \\
\hline
\end{tabular}
}
\end{center}
\vskip -0.2in
\end{table}

%
{\bf Permutation-sensitive TGNNs.}
There are several temporal graph works that consider permutation-sensitive
(non-equivariant) graph representations.
DynAERNN~\citep{goyal2020dyngraph2vec} uses adjacency vector of each node as
augmented node feature; AdaNN-D~\citep{xu2019adaptive} uses the random walk
vector based on adjacency matrix as augmented node feature; E-LSTM-D~\citep{%
    chen2019e%
} uses flattened adjacency matrix as augmented input of LSTM; and
GC-LSTM~\citep{chen2018gc} directly uses adjacency matrix as augmented input of
LSTM.
All aforementioned works share the same challenge:
If we permute node indices, resulting in a different adjacency matrix of an
isomorphic graph, their final node representation changes.
This is a property we want to avoid in node and graph tasks (classification and
regression).

%
{\bf Extension of Temporal Graph Representations.}
Recent works also merge existing TGNNs with new representation tuning
techniques.
\citet{hajiramezanali2019variational} incorporates variable autoencoder into
\timeandgraph.
\citet{lei2019gcn,xiong2019dyngraphgan} incorporates \timeandgraph under an
adversarial learning framework.
Similarly, HTGN~\citep{yang2021discrete} is a \graphthentime representation in
hyperbolic space.
Although these variants provide new applications for temporal graph
representations, they do not improve the expressivity of the \timeandgraph
framework.

\begin{table}[t]
\caption{
    {\bf Performance on temporal graph and node classification tasks.}
    Performance is evaluated by ROCAUC score where higher value means better
    performance.
    Best mean performance is both bolded and underlined, and the second best
    one is only bolded.
    Our proposed GCN-GRU is the only one works on DynCSL dataset, and is placed
    at the top with another \timethengraph model, TGN, on Brain10 dataset.
}
\label{tab:classification}
%
\begin{center}
\resizebox{\columnwidth}{!}{
\begin{tabular}{clrr}
\hline
Representation &
\multicolumn{1}{c}{Model} & \multicolumn{1}{c}{DynCSL} &
\multicolumn{1}{c}{Brain-10} \\
\hline
\multirow{4}{*}{\graphthentime} & EvolveGCN-O &
$0.50 {\scriptstyle \pm 0.00}$ &
$0.58 {\scriptstyle \pm 0.10}$ \\
& EvolveGCN-H & $0.50 {\scriptstyle \pm 0.00}$ &
$0.60 {\scriptstyle \pm 0.11}$ \\
& GCN-GRU & $0.50 {\scriptstyle \pm 0.00}$ & $0.87 {\scriptstyle \pm 0.07}$ \\
& DySAT & $0.50 {\scriptstyle \pm 0.00}$ & $0.77 {\scriptstyle \pm 0.07}$ \\
\hline
\multirow{2}{*}{\timeandgraph} & GCRN-M2 & $0.52 {\scriptstyle \pm 0.04}$ &
$0.77 {\scriptstyle \pm 0.04}$ \\
& DCRNN & $0.51 {\scriptstyle \pm 0.03}$ & $0.84 {\scriptstyle \pm 0.02}$ \\
\hline
\multirow{3}{*}{\timethengraph} & TGAT & $0.48 {\scriptstyle \pm 0.03}$ &
$0.80 {\scriptstyle \pm 0.03}$ \\
& TGN & $0.51 {\scriptstyle \pm 0.04}$ &
\underline{$\mathbf{0.91 {\scriptstyle \pm 0.03}}$} \\
& GRU-GCN & \underline{$\mathbf{1.00 {\scriptstyle \pm 0.00}}$} &
\underline{$\mathbf{0.91 {\scriptstyle \pm 0.03}}$} \\
\hline
\end{tabular}
}
\end{center}
\vskip -0.2in
\end{table}

\begin{table*}[t]
\caption{
    {\bf Performance on temporal node regression tasks.}
    Performance is evaluated by MAPE (\%) where lower value means better
    performance (see \Cref{eqn:mape} in \Cref{subsec:hypers}).
    Best mean performance is both bolded and underlined, and the second best
    one is only bolded.
    Our proposal GCN-GRU is always the best one on all inductive learnings.
    On transductive tasks, our proposal is the best on PeMS08 and England-COVID
    datasets, and is the second best on PeMS04 and Spain-COVID datasets.
    \Timethengraph representations can always provide the best performance with
    only one exception for transductive learning on PeMS04 where it is only the
    second best.
}
\label{tab:regression}
%
\begin{center}
\resizebox{\textwidth}{!}{
\begin{tabular}{clrrrrrrrr}
\hline
\multirow{2}{*}{Representation} &
\multicolumn{1}{c}{\multirow{2}{*}{Model}} &
\multicolumn{2}{c}{PeMS04} & \multicolumn{2}{c}{PeMS08} &
\multicolumn{2}{c}{Spain-COVID} & \multicolumn{2}{c}{England-COVID} \\
& & \multicolumn{1}{c}{Transductive} & \multicolumn{1}{c}{Inductive} &
\multicolumn{1}{c}{Transductive} & \multicolumn{1}{c}{Inductive} &
\multicolumn{1}{c}{Transductive} & \multicolumn{1}{c}{Inductive} &
\multicolumn{1}{c}{Transductive} & \multicolumn{1}{c}{Inductive} \\
\hline
\multirow{4}{*}{\graphthentime} & EvolveGCN-O &
$3.20 {\scriptstyle \pm 0.25}$\% & $2.61 {\scriptstyle \pm 0.42}$\% &
$2.65 {\scriptstyle \pm 0.12}$\% & $2.40 {\scriptstyle \pm 0.27}$\% &
$2.64 {\scriptstyle \pm 0.12}$\% & $2.02 {\scriptstyle \pm 0.11}$\% &
$4.07 {\scriptstyle \pm 0.73}$\% & $3.88 {\scriptstyle \pm 0.47}$\% \\
& EvolveGCN-H &
$3.34 {\scriptstyle \pm 0.14}$\% & $2.84 {\scriptstyle \pm 0.31}$\% &
$2.81 {\scriptstyle \pm 0.28}$\% & $2.81 {\scriptstyle \pm 0.23}$\% &
$2.62 {\scriptstyle \pm 0.33}$\% & $2.09 {\scriptstyle \pm 0.30}$\% &
$4.14 {\scriptstyle \pm 1.14}$\% & $3.50 {\scriptstyle \pm 0.42}$\% \\
& GCN-GRU &
\underline{$\mathbf{1.60 {\scriptstyle \pm 0.14}}$}\% &
$1.28 {\scriptstyle \pm 0.04}$\% &
$1.40 {\scriptstyle \pm 0.26}$\% & $1.07 {\scriptstyle \pm 0.03}$\% &
$2.39 {\scriptstyle \pm 0.06}$\% & $1.22 {\scriptstyle \pm 0.66}$\% &
$\mathbf{3.56 {\scriptstyle \pm 0.26}}$\% &
$\mathbf{2.97 {\scriptstyle \pm 0.34}}$\% \\
& DySAT &
$1.86 {\scriptstyle \pm 0.08}$\% & $1.58 {\scriptstyle \pm 0.08}$\% &
$1.49 {\scriptstyle \pm 0.08}$\% & $1.34 {\scriptstyle \pm 0.03}$\% &
$2.15 {\scriptstyle \pm 0.18}$\% &
$\mathbf{0.89 {\scriptstyle \pm 0.44}}$\% & $3.67 {\scriptstyle \pm 0.15}$\% &
$3.32 {\scriptstyle \pm 0.76}$\% \\
\hline
\multirow{2}{*}{\timeandgraph} & GCRN-M2 &
$1.70 {\scriptstyle \pm 0.20}$\% & $1.20 {\scriptstyle \pm 0.06}$\% &
$\mathbf{1.30 {\scriptstyle \pm 0.17}}$\% & $1.00 {\scriptstyle \pm 0.10}$\% &
$1.94 {\scriptstyle \pm 0.54}$\% & $1.54 {\scriptstyle \pm 0.50}$\% &
$3.85 {\scriptstyle \pm 0.39}$\% & $3.37 {\scriptstyle \pm 0.27}$\% \\
& DCRNN &
$1.67 {\scriptstyle \pm 0.19}$\% & $1.27 {\scriptstyle \pm 0.06}$\% &
$1.32 {\scriptstyle \pm 0.19}$\% & $1.07 {\scriptstyle \pm 0.03}$\% &
$2.12 {\scriptstyle \pm 0.33}$\% & $0.90 {\scriptstyle \pm 0.21}$\% &
$3.58 {\scriptstyle \pm 0.53}$\% & $3.09 {\scriptstyle \pm 0.24}$\% \\
\hline
\multirow{3}{*}{\timethengraph} & TGAT &
$3.11 {\scriptstyle \pm 0.50}$\% & $2.25 {\scriptstyle \pm 0.27}$\% &
$2.66 {\scriptstyle \pm 0.27}$\% & $2.34 {\scriptstyle \pm 0.19}$\% &
$2.46 {\scriptstyle \pm 0.04}$\% & $1.81 {\scriptstyle \pm 0.14}$\% &
$5.44 {\scriptstyle \pm 0.46}$\% & $5.13 {\scriptstyle \pm 0.26}$\% \\
& TGN &
$1.79 {\scriptstyle \pm 0.21}$\% & $\mathbf{1.19 {\scriptstyle \pm 0.07}}$\% &
$1.49 {\scriptstyle \pm 0.26}$\% & $\mathbf{0.99 {\scriptstyle \pm 0.06}}$\% &
\underline{$\mathbf{1.62 {\scriptstyle \pm 0.33}}$}\% &
$1.25 {\scriptstyle \pm 0.48}$\% & $4.15 {\scriptstyle \pm 0.81}$\% &
$3.17 {\scriptstyle \pm 0.23}$\% \\
& GRU-GCN &
$\mathbf{1.61 {\scriptstyle \pm 0.35}}$\% &
\underline{$\mathbf{1.13 {\scriptstyle \pm 0.05}}$}\% &
\underline{$\mathbf{1.27 {\scriptstyle \pm 0.21}}$}\% &
\underline{$\mathbf{0.89 {\scriptstyle \pm 0.07}}$}\% &
$\mathbf{1.66 {\scriptstyle \pm 0.63}}$\% &
\underline{$\mathbf{0.65 {\scriptstyle \pm 0.16}}$}\% &
\underline{$\mathbf{3.41 {\scriptstyle \pm 0.28}}$}\% &
\underline{$\mathbf{2.87 {\scriptstyle \pm 0.19}}$}\% \\
\hline
\end{tabular}
}
\end{center}
\vskip -0.2in
\end{table*}

\section{Experiments}
\label{sec:experiments}

In this section, we evaluate a simple architecture based on our \timethengraph
framework on one synthetic dataset and five different real-world datasets.
We also evaluate eight state-of-the-art temporal graph representation baselines
on the same tasks.
Each experiment is repeated ten times with different random initialization.
Please refer to the \Cref{sec:configs} for a detailed description of experiment
setup, and to our code for hyperparameter configurations~\footnote{
    \href{
        https://www.dropbox.com/s/gqtzpngj5vaah9s/ICML2022-Code-Submit.zip?dl=0
    }{Source code}.
}.

{\bf Learning tasks.}
In this work, all the learning tasks will use equivariant temporal graph
representations as described in \Cref{sec:express}.
It is known that equivariant graph representations are generally not
sufficiently expressive for prediction tasks beyond node and graph (structure)
predictions, especially for edge-level tasks such as link prediction~\citep{%
    srinivasan2020on,you2019position%
} and shortest path estimation~\citep{%
    dehmamy2019understanding,loukas2020what,tang2020towards%
}.
Indeed, {\em equivariant} temporal graph representations inherit the same
expressivity insufficiency, thus are also improper for edge-level tasks.
A formal explanation of this limitation is provided in \Cref{%
    subsec:proof_link_pred%
}.
Thus, to deliver theoretically sound experiments, we only consider
temporal graph classification, node classification and node regression as
learning tasks in our experiments.

{\bf DynCSL.}
DynCSL is a synthetic temporal graph classification task.
Each sample is constructed by a sequence of 8 graph snapshots.
Each snapshot is randomly drawed from $
    \big\{ \gpC_{19, 2}, \cdots, \gpC_{19, 6} \big\}
$.
Here, $\gpC_{|\stV|, s}$ means a Circular Skip Link (CSL)~\citep{%
    murphy2019relational%
} graph with $|\stV|$ nodes and skip length $s$.
A CSL sequence example is introduced \Cref{fig:dyncsl}.
The goal is to predict the number of unique non-isomorphic CSLs in the
constructed temporal graph, e.g., the top temporal graph in \Cref{fig:dyncsl}
has prediction ``2'', while the bottom one has prediction ``1''.
DynCSL is a temporal extension of the CSL task. CSL is widely used as a first-test of the expressiveness of static GNNs~\citep{bodnar2021weisfeiler,chen2019equivalence,cotta2021reconstruction,dwivedi2020benchmarking,nguyen2020graph,tiezzi2021deep,zhang2021eigen}.

{\bf Real-world applications.}
We also evaluate the performace on several real-word applcations including
a brain functionality classification Brain10, two traffic forecasting PeMS04
and PeMS08, and two COVID spreading prediction Spain-COVID and England-COVID.

Generally speaking, our data consists of a set of temporal graphs.
These temporal graphs are defined over the same set of nodes.
For instance, a temporal graph may represent the evolution of a system in a
day, while the different temporal graphs would represent different days.
Please refer to \Cref{subsec:datasets} for a in-depth description of our
datasets, whose general statistics are shown in \Cref{tab:meta}.

{\bf Our model and baselines.}
We compare an representation instance of our \timethengraph approach, GRU-GCN,
with 5 \timeandgraph methods (namely, EvolveGCNs, GCN-GRU, DySAT, GCRN-M2 and
DCRNN) and two \timethengraph methods (namely, TGAT and TGN).
Our proposal GRU-GCN means using GRUs for both node and edge sequence
representations and GCN as graph representation in \timethengraph framework
as \Cref{eqn:timethengraph}.
All baselines have been briefly introduced in \Cref{sec:related}.
Please refer to \Cref{sec:configs} for details about architectures of all
models.

\begin{table*}[t]
\caption{
    {\bf Resource cost of tasks on GPUs.}
    We collect the peak GPU memory and average training time per minibatch from
    all tasks utilizing GPU resources on a GeForce RTX 2080 Ti.
    Best training time efficency is both bolded and underlined.
    Our proposal GCN-GRU are always the most time efficent method for GPUs on
    all four tasks.
    Furthermore, it uses the least GPU memory among all state-of-the-art
    \timethengraph representation methods.
}
\label{tab:rescost}
%
\begin{center}
\resizebox{\textwidth}{!}{
\begin{tabular}{clrrrrrrrr}
\hline
\multirow{3}{*}{Representation} &
\multicolumn{1}{c}{\multirow{3}{*}{Model}} &
\multicolumn{2}{c}{PeMS04} & \multicolumn{2}{c}{PeMS08} &
\multicolumn{2}{c}{Spain-COVID} & \multicolumn{2}{c}{England-COVID} \\
& & \multicolumn{1}{c}{\multirow{1}{*}{\makecell{Peak GPU\\Memory}}} &
\multicolumn{1}{c}{
    \multirow{1}{*}{\makecell{Average Training\\Time per Minibatch}}
} & \multicolumn{1}{c}{\multirow{1}{*}{\makecell{Peak GPU\\Memory}}} &
\multicolumn{1}{c}{
    \multirow{1}{*}{\makecell{Average Training\\Time per Minibatch}}
} & \multicolumn{1}{c}{\multirow{1}{*}{\makecell{Peak GPU\\Memory}}} &
\multicolumn{1}{c}{\multirow{1}{*}{
    \makecell{Average Training\\Time per Minibatch}}
} & \multicolumn{1}{c}{\multirow{1}{*}{\makecell{Peak GPU\\Memory}}} &
\multicolumn{1}{c}{
    \multirow{1}{*}{\makecell{Average Training\\Time per Minibatch}}
} \\
& & & & & & & & & \\
\hline
\multirow{4}{*}{\graphthentime} & EvolveGCN-O & 86 MB & 19ms & 55 MB & 17 ms &
221 MB & 14 ms & 3MB & 9 ms \\
& EvolveGCN-H & 205 MB & 40 ms & 130 MB & 31 ms & 512 MB & 21 ms & 4 MB &
15 ms \\
& GCN-GRU & 1089 MB & 17 ms & 602 MB & 15 ms & 140 MB & 12 ms & 6 MB & 8 ms \\
& DySAT & 1911 MB & 26 ms & 1060 MB & 24 ms & 137 MB & 18 ms & 7 MB & 14 ms \\
\hline
\multirow{2}{*}{\timeandgraph} & GCRN-M2 & 3099 MB & 195 ms & 1871 MB &
159 ms & 5423 MB & 124 ms & 22 MB & 84 ms \\
& DCRNN & 1730 MB & 83 ms & 1024 MB & 65 ms & 2460 MB & 50 ms & 13 MB &
34 ms \\
\hline
\multirow{3}{*}{\timethengraph} & TGAT & 7945 MB & 101 ms & 5680 MB & 72 ms &
7300 MB & 94 ms & 96 MB & 21 ms \\
& TGN & 3963 MB & 25 ms & 2908 MB & 19 ms & 5205 MB & 29 ms & 73 MB & 16 ms \\
& GRU-GCN & 859 MB & \underline{\bf 7 ms} & 574 MB & \underline{\bf 5 ms} &
1538 MB & \underline{\bf 10 ms} & 52 MB & \underline{\bf 3 ms} \\
\hline
\end{tabular}
}
\end{center}
\vskip -0.2in
\end{table*}

\subsection{Performance on tasks that mostly depend on the evolving topology}
\label{subsec:result-struct}

First, we analyze the performance of our model against other baselines on
DynCSL and Brain10 datasets whose prediction are highly dependent to temporal
topologies in \Cref{tab:classification}.
DynCSL is an extension of the task in the proof of \Cref{thm:temp_1wl_express}
where our method is strictly more expressive than \timeandgraph on classifying
temporal topologies.
For Brain10, the functionality of brain voxel is mostly determined by its
activations with other voxels which are mostly covered in its temporal
topologies, thus we also expect our proposal to perform better than
\timeandgraph baselines on it.

Indeed, our GRU-GCN \timethengraph architecture is the only method that works
on the DynCSL task, while all the other baselines ---including two
state-of-the-art \timethengraph methods--- give performances no better than
random predictions.
This observation follows the \timethengraph expressivity advantage introduced in
\Cref{fig:dyncsl}.
Notably, the result also reflects our proposed GRU-GCN also holds expressivity
advantage over the other two \timethengraph methods, TGAT and TGN.
This advantage is caused by taking not-existing edge attributes into
consideration in our model while TGAT and TGN only collects existing edge
attributes.
Please refer to \Cref{subsec:tgn_on_dyncsl} for formal explanation of this
observation.

Our \timethengraph GRU-GCN is also one of the top methods on Brain10, with TGN
as the only competitor of the same performance.
These two are both under \timethengraph framework, and gain a clear improvement
from \timeandgraph baselines.
This observation provides a further evidence on expressivity advantage of
\timethengraph over \timeandgraph.
The exception is TGAT, which is not performing ideally in both cases.
The reason is that there is no sequence representation used in TGAT which
plays a key role in \timethengraph framework, thus TGAT is more like a static
GNN baseline but on heterogeneous graph rather than a temporal graph method
which limits its power compared to other temporal graph representations.

\subsection{Performance on tasks where attribute evolution is more relevant}
\label{subsec:result-other}

\Cref{tab:classification} summarizes the result over all remaining real-world
datasets.
Each dataset is further split into two different tasks:
{\em Transductive learning}, where two disjoint sets of nodes of a temporal
graph are selected for validation and test.
At the last snapshot, the attributes of nodes in validation and test data
are hidden from us during training.
{\em Inductive learning}, where in training we have access to full data (all
node and edges attributes over all snapshots) for some temporal graphs
(training graphs).
In test, for a new set of unseen temporal graphs, we are asked to predict all
node attributes of the last snapshot given past snapshots.

\Cref{tab:regression} shows that our proposed \timethengraph GRU-GCN is often
the best model on all tasks (tansductive and inductive), with only two
exceptions:
Transductive PeMS04 and Spain-COVID where our proposal is still the second best
model and very close to the best performance.
This is somewhat expected since the rich variation on real-world node and edge
attributes in these datasets can reduce the expressivity advantage of
\timethengraph over \timeandgraph.

\subsection{Empirical computational efficiency}
\label{subsec:result-cost}

While theoretically we do not expect computational efficiency gains from
\timethengraph methods over \timeandgraph methods (see \Cref{%
    subsec:complexity%
}), in practice we observe reasonable gains in some datasets.
In order to study the real-world computational efficiency of our approach on
GPUs, we collect the peak GPU memory cost and average training time cost per
minibatch for datasets utilizing GPU resources.
We do not collect GPU data for Brain10 since its graph is too large to fit in
GPU memory for our comparisons.

The runtime data in \Cref{tab:rescost} shows that our GRU-GCN proposal is
consistently the fastest method in all datasets.
Furthermore, it also requires the least amount of memory among all
\timethengraph methods.
This efficiency advantage is caused by small number of edges for
studying datasets, which results in small cost for RNN on edge attributes.
However, if the aggregated temporal graph is dense,
\timethengraph will no longer be as efficient.
We provide more details in \Cref{subsec:rescost}.

\section{Conclusion and Future Work}
\label{sec:conclusion}
We proposed \timethengraph, an alternative representation framework for
temporal graphs for node and graph classification and regression tasks.
While \timeandgraph (and its variant \graphthentime) are still the most
widely-used frameworks for temporal graphs, we formally prove that our proposed
\timethengraph framework is more expressive than \timeandgraph if the graph
representation uses the ubiquitous 1-WL GNN architecture.
Our experiments confirm this expressivity gain, achieving state-of-the-art
performance on synthetic and real-world tasks.
Overall, \timethengraph provides a new ML tool for temporal graph applications.

\pagebreak
\bibliography{references}
\bibliographystyle{icml2022}

\newpage
\appendix
\onecolumn
\section{Proves}
\label{sec:proves}

\subsection{Identifiable Set Equivalence}
\label{subsec:proof_biject}

\begin{lemma}[Identifiable Set Equivalence]
\label{lem:biject}
Suppose $\stI_{1}(f, \stU)$ and $\stI_{2}(f, \stU)$ are two identifiable sets
of representation function $f$ on domain $\stU$, then there exists a bijection
between $\stI_{1}(f, \stU)$ and $\stI_{2}(f, \stU)$ based on matching between
representations.
\end{lemma}

\begin{proof}

We show this by contradiction.
Assume that there is no such bijection by matching representations between
$\stI_{1}( f, \stU)$ and $\stI_{2}(f, \stU)$.
Without loss of generality, suppose there is an element $
    \vcu_{1} \in \stI_{1}(f, \stU)
$ such that no element from $\stI_{2}(f, \stU)$ can match its representation,
in other words, $
    \exists \vcu_{1} \in \stI_{1}(f, \stU), \forall \vcu_{2} \in \stI_{2}(
        f, \stU
    ), f\big( \vcu_{1} \big) \neq f\big( \vcu_{2} \big)
$.
This implies $\vcu_{1} \notin \stI_{2}(f, \stU)$,
otherwise $\vcu_{1} \in \stI_{2}(f, \stU)$ will match with $
    \vcu_{1} \in \stI_{1}(f, \stU)
$ on representations.
Thus, $\vcu_{1} \in \stU - \stI_{2}(f, \stU)$.
According to \Cref{def:identify_set}, for $
    \vcu_{1} \in \stU - \stI_{2}(f, \stU)
$, $
    \exists \vcu'_{2} \in \stI_{2}(f, \stU), f\big( \vcu_{1} \big) = f\big(
    \vcu'_{2} \big)
$.
This leads to a contradiction since we assume that no elements from $
    \stI_{2}(f, \stU)
$ can match with $\vcu_{1}$ on its representation $f\big( \vcu_{1} \big)$, but
we find matching $\vcu'_{2} \in \stI_{2}(f, \stU)$.

So, for any elements in $\stI_{1}(f, \stU)$, there must exists an element in
$\stI_{2}(f, \stU)$ matching its representation.
And, this mapping is an injection from $\stI_{1}(f, \stU)$ to $
    \stI_{2}(f, \stU)
$ since $\stI_{1}(f, \stU)$ does not contain elements of the same
representation based on \Cref{def:identify_set}.
The same applies flipping the roles of  $\stI_{2}(f, \stU)$ and $
    \stI_{1}(f, \stU)
$.

Hence, by the Schröder-Bernstein theorem, if there is an injection from
$\stI_{1}(f, \stU)$ to $\stI_{2}(f, \stU)$ and an injection from $
    \stI_{2}(f, \stU)
$ to $\stI_{1}(f, \stU)$, then there is a bijection between $\stI_{1}(f, \stU)$
and $\stI_{2}(f, \stU)$.

\end{proof}

\subsection{Temporal 1-WL Expressivity}
\label{subsec:proof_temp_1wl_express}

\timeonewl*

\begin{proof}

{\bf Roadmap:}
We first show that we can construct an equivalent \timethenonewl representation
for any \timeandonewl representation on arbitrary temporal graph domain $
    \fmT_{|\stV|, T, p}
$.
Thus, according to \Cref{lem:express_more_alter}, \timethenonewl is at least as
expressive as \timeandonewl.
Furthermore, we provide an instance where \timeandonewl fails to distinguish
two different temporal graphs but \timethenonewl can.
This further shows that \timethenonewl is strictly more expressive, in other
words,
\begin{align*}
\text{\Timeandonewl} \exprlt_{\fmT_{|\stV|, T, p}} \text{\Timethenonewl}.
\end{align*}

\begin{figure*}[p]
\vskip -0.2in
\centering
\begin{minipage}[c]{0.48\textwidth}
\centering
\includegraphics[width=0.95\linewidth]{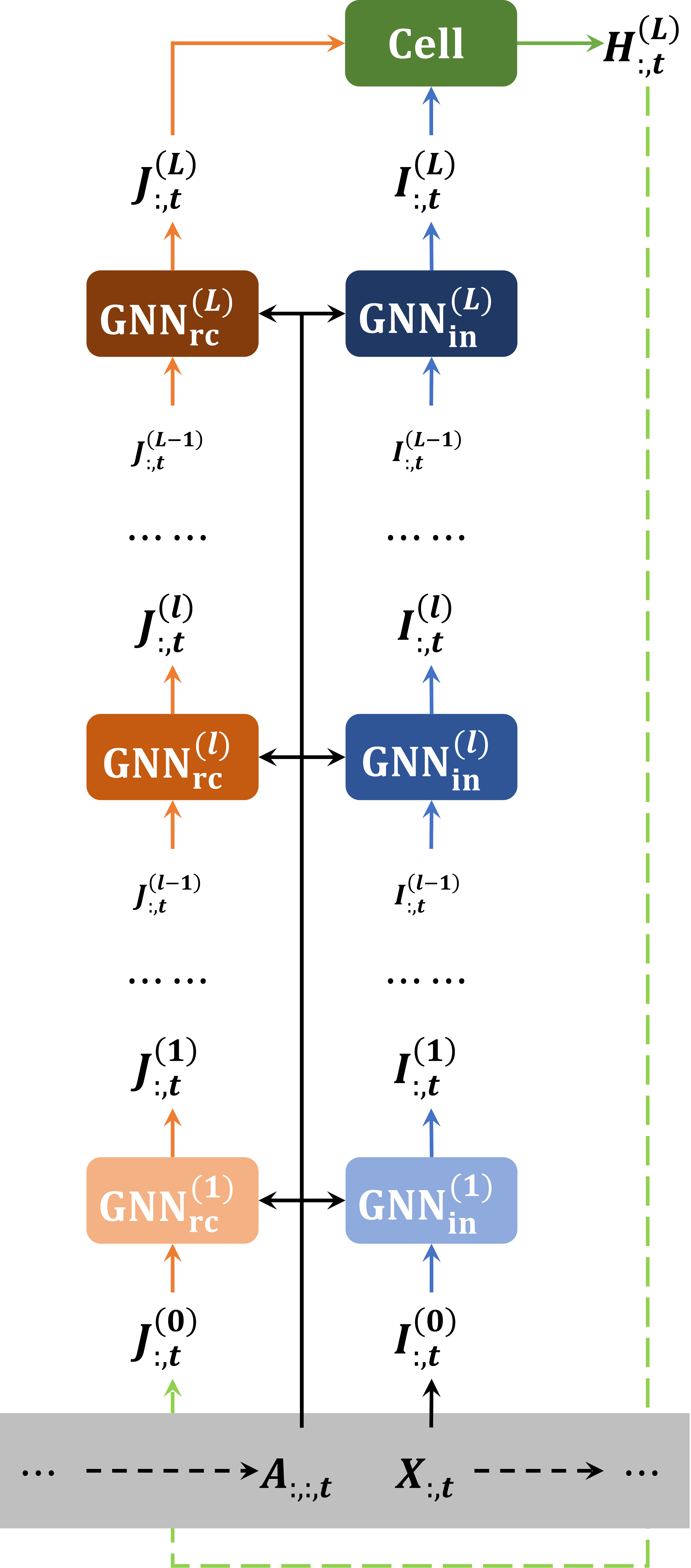}
\caption{
    {\bf Illustration of a \timeandgraph architecture:}
    $\tnX_{:, t}$ is all node attributes and $\tnA_{:, :, t}$ is all edge
    attributes at snapshot $t$ for $1 \leq t \leq T$.
    $\text{GNN}_\text{in}^{(l)}$ (blue) and $\text{GNN}_\text{rc}^{(l)}$
    (orange) are $l$-th layer of $\text{GNN}_\text{in}^{L}$ and $
        \text{GNN}_\text{rc}^{L}
    $ as defined in \Cref{eqn:timeandgraph}.
    Different color saturation means layer depth.
    $\text{Cell}$ (green) is the RNN cell defined in \Cref{eqn:timeandgraph}.
    $\tnH_{:, t}$ is the historical representation output of \timeandgraph
    until snapshot $t$, and $\tnH_{:, T}$ will be the final representation
    output.
    $\tnI_{:, t}^{(l)}$ is the internal (hidden) states of $
        \text{GNN}_\text{in}^{L}
    $ at layer $l$.
    $\tnJ_{:, t}^{(l)}$ is the internal (hidden) states of $
        \text{GNN}_\text{rc}^{L}
    $ at layer $l$.
}
\label{fig:time_and_graph}
\end{minipage}
\hfill
\begin{minipage}[c]{0.48\textwidth}
\centering
\includegraphics[width=0.95\linewidth]{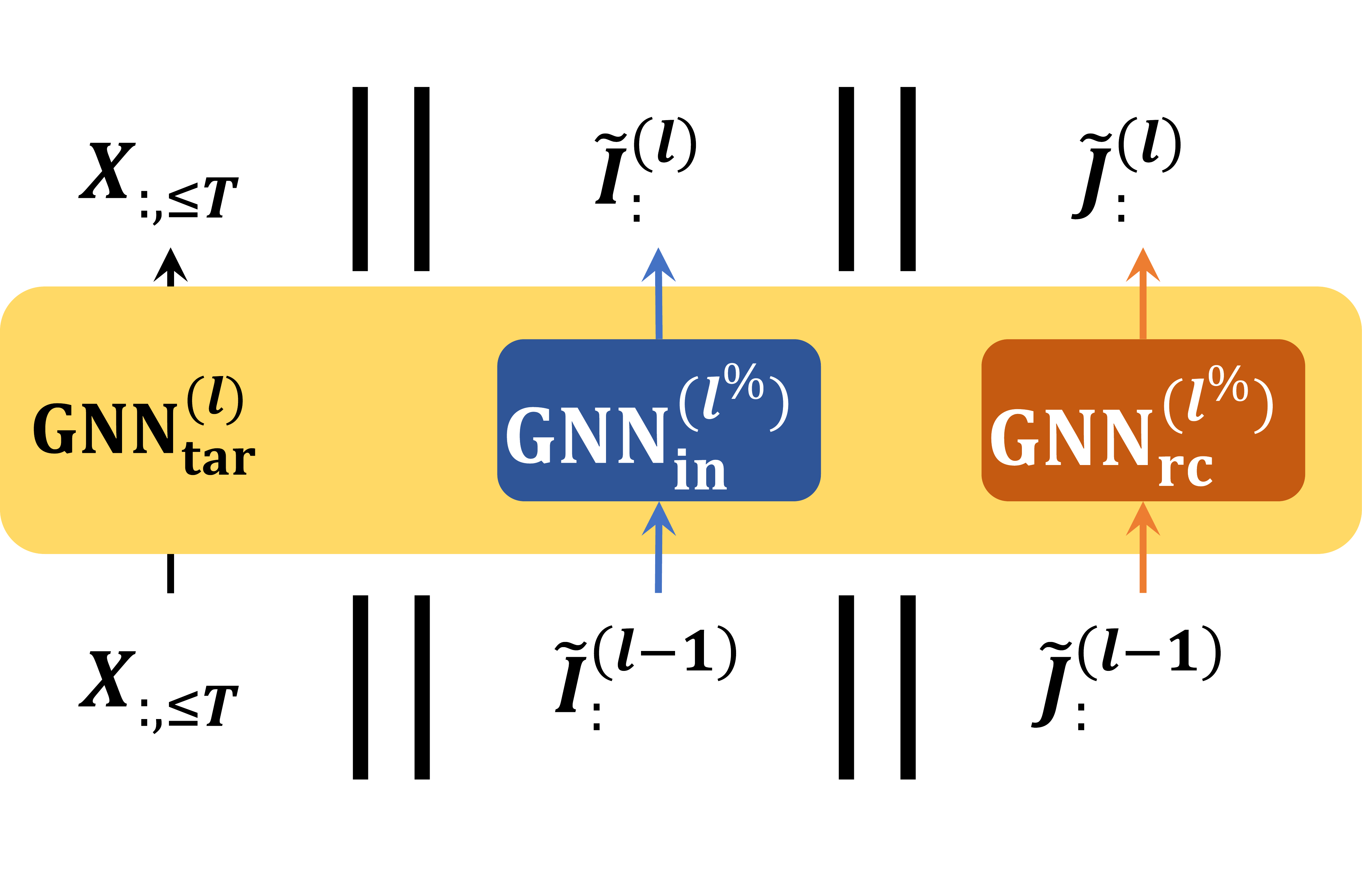}
\caption{
    {
        \bf Illustration of $l$-th constructed \timethengraph layer in $
            \mathbf{\text{GNN}_\text{tar}^{TL}}
        $ when \underline{$\mathbf{l \text{ mod } L \neq 0}$}:
    }
    Yellow square is the illustration of \Cref{%
        eqn:construct_msg,eqn:construct_update%
    } when $t = \lceil l / L \rceil$.
    Operations (blue and orange squares) inside it imply the same operations as
    in \Cref{fig:time_and_graph} with the same colors (regardless of
    saturation) and superscript values.
    We can see that for $l \text{ mod } L \neq 0$, any arbitrary layer $l$ of
    the $\text{GNN}_\text{tar}^{TL}$ is indeed processing layer $l^{\%}$ of $
        \text{GNN}_\text{in}^{L}
    $ and $\text{GNN}_\text{rc}^{L}$ at snapshot $t = \lceil l / L \rceil$ in
    parallel as \Cref{fig:time_and_graph}.
}
\label{fig:layer_mid}
\vfill
\includegraphics[width=0.95\linewidth]{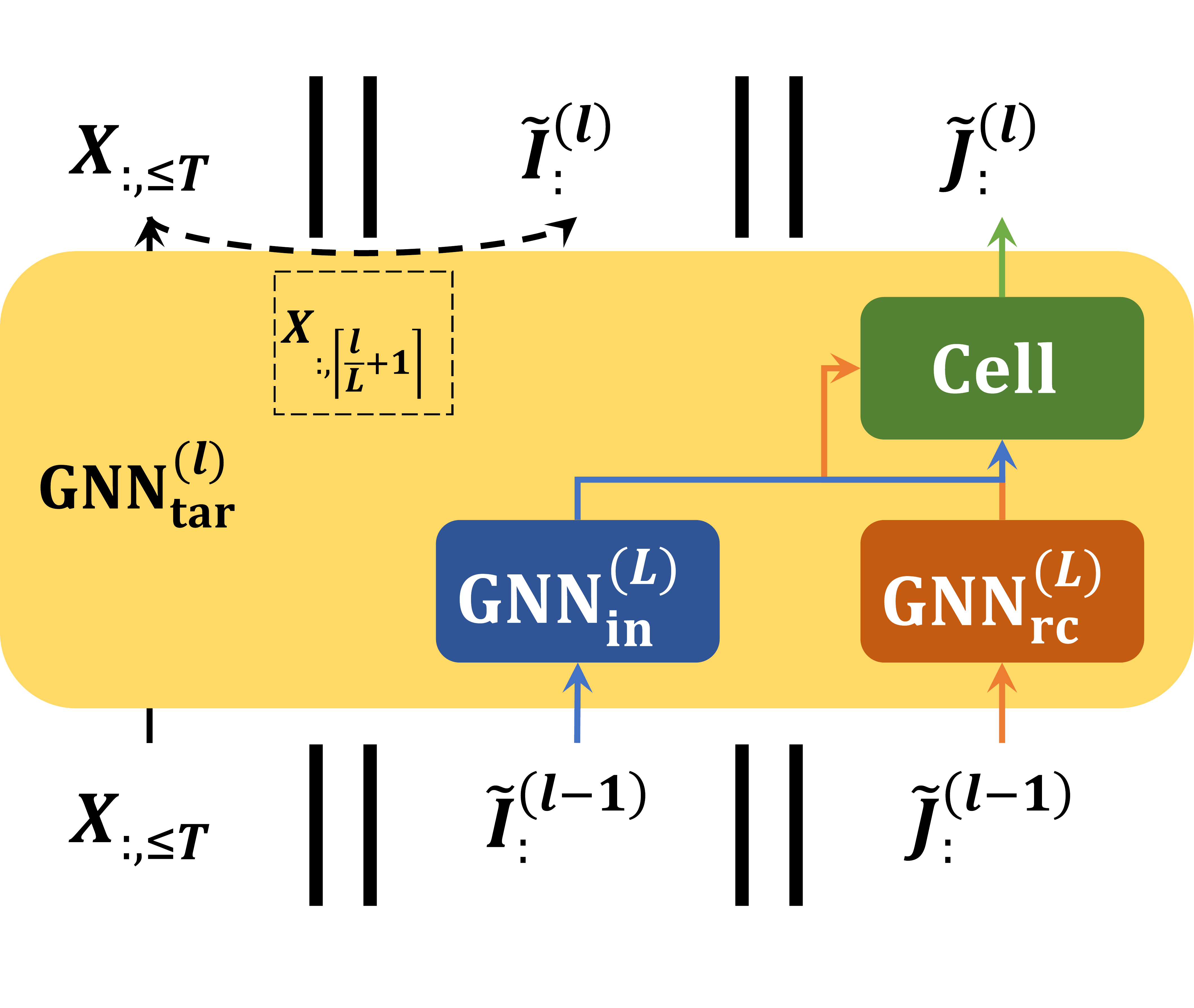}
\caption{
    {
        \bf Illustration of $l$-th constructed \timethengraph layer in $
            \mathbf{\text{GNN}_\text{tar}^{TL}}
        $ when \underline{$\mathbf{l \text{ mod } L = 0}$}:
    }
    Yellow square is the illustration of \Cref{%
        eqn:construct_msg,eqn:construct_update%
    } when $t = \lceil l / L \rceil$.
    Operations (green, blue and orange squares) inside it imply the same
    operations as in \Cref{fig:time_and_graph} with the same colors (regardless
    of saturation) and superscript values.
    We can see that for $l \text{ mod } L = 0$, different from the operations
    illustrated in the \Cref{fig:layer_mid}, $\text{GNN}_\text{tar}^{(l)}$ will
    replace part of output, $\bld{\tnI}_{:}^{l}$, by node attributes $
        \tnX_{:, \lceil l / L + 1 \rceil}
    $, and will replace the other part of output, $\bld{\tnJ}_{:}^{l}$, by the
    output of \Cref{eqn:timeandgraph_cell} which indeed is mirroring the
    operations after layer $L$ in \Cref{fig:time_and_graph} (layer $L$ itself
    and recurrent cell).
}
\label{fig:layer_fin}
\end{minipage}
\vskip -0.2in
\end{figure*}

{
    \bf First, we will show that any \timeandgraph representation function has
    an equivalent \timethengraph function.
}
We start with recalling the definition of \timeandgraph given by
\Cref{eqn:timeandgraph}:
\begin{align*}
\tnH_{i, t} = \text{Cell}\bigg(
    \Big[
        \text{GNN}_\text{in}^{L}\big( \tnX_{:, t}, \tnA_{:, :, t} \big)
    \Big]_{i},
    \Big[
        \text{GNN}_\text{rc}^{L}\big( \tnH_{:, t - 1}, \tnA_{:, :, t} \big)
    \Big]_{i}
\bigg)
\end{align*}
The above definition has two different graph neural networks $
    \text{GNN}_\text{in}^{L}
$ and $\text{GNN}_\text{rc}^{L}$.
These represent encoding raw node attributes at time $t$ and encoding node
historical representations before $t$ along with topology $\tnA_{:, :, t}$ at
snapshot $t$, respectively.

Also, we revisit the definition of GNN in \Cref{eqn:gnn} of any arbitrary layer
$l$ for the sake of later construction:
\begin{equation*}
\begin{aligned}
\tnM_{i}^{(l)} & = \sum\limits_{j \in \stN(i)} \text{MSG}^{(l)}\big(
    \tnZ_{i}^{(l - 1)}, \tnZ_{j}^{(l - 1)}, \tnA_{i, j}
\big),
\\
\tnZ_{i}^{(l)} & = \text{UPDATE}^{(l)}\big(
    \tnZ_{i}^{(l - 1)}, \tnM_{i}^{(l)}
\big),
\end{aligned}
\end{equation*}

The full \timeandgraph architecture is shown in \Cref{fig:time_and_graph} and
is formally described by 3 parts:
\begin{enumerate}
\item
{\bf Description of $\text{GNN}_\text{in}^{L}$:}
This GNN embeds the input features at time $t \in \{1, \cdots, T\}$.
To differentiate symbols from other GNN operations, we denote the $
    \text{GNN}_\text{in}^{(l)}
$ definition at each layer $l \in \{1, \cdots, L\}$ as
\begin{equation}
\label{eqn:timeandgraph_in}
\begin{aligned}
\tnE_{i, t}^{(l)} & = \sum\limits_{j \in \stN_{t}(i)}
\text{MSG}_\text{in}^{(l)} \big(
    \tnI_{i, t}^{(l - 1)}, \tnI_{j, t}^{(l - 1)}, \tnA_{i, j, t}
\big),
\\
\tnI_{i, t}^{(l)} & = \text{UPDATE}_\text{in}^{(l)}\big(
    \tnI_{i, t}^{(l - 1)}, \tnE_{i, t}^{(l)}
\big),
\end{aligned}
\end{equation}
where $\tnI_{i, t}^{(l)}$ represents the embedding of arbitrary node $
    i \in \stV
$ obtained at layer $l$ on snapshot $t$ and $\stN_{t}(i)$ defines the neighbor
set of $i$ on snapshot $t$.
We also initialize the corner case $\tnI_{i, t}^{(0)} = \tnX_{i, t}$.
$\text{MSG}_\text{in}^{(l)}$ and $\text{UPDATE}_\text{in}^{(l)}$ are learnable
parametric functions of arbitrary forms, e.g., Multi-Layer Perceptions (MLPs).
Finally, the output for arbitrary node $i \in \stV$ is $
    \big[
        \text{GNN}_\text{in}^{L}\big( \tnX_{:, t}, \tnA_{:, :, t} \big)
    \big]_{i} := \tnI_{i, t}^{(L)}
$.
\item
{\bf Description of $\text{GNN}_\text{rc}^{L}$:}
We denote the $\text{GNN}_\text{rc}^{(l)}$ definition at each layer $
    l \in \{1, \cdots, L\}
$ as
\begin{equation}
\label{eqn:timeandgraph_rc}
\begin{aligned}
\tnF_{i, t}^{(l)} & = \sum\limits_{j \in \stN_{t}(i)}
\text{MSG}_\text{rc}^{(l)} \big(
    \tnJ_{i, t}^{(l - 1)}, \tnJ_{j, t}^{(l - 1)}, \tnA_{i, j, t}
\big),
\\
\tnJ_{i, t}^{(l)} & = \text{UPDATE}_\text{rc}^{(l)}\big(
    \tnJ_{i, t}^{(l - 1)}, \tnF_{i, t}^{(l)}
\big),
\end{aligned}
\end{equation}
where $\tnJ_{i, t}^{(l)}$ represents the embedding of arbitrary node $
    i \in \stV
$ obtained at layer $l$ on snapshot $t$ and $\stN_{t}(i)$ defines the neighbor
set of $i$ on snapshot $t$.
We also initialize the corner case $\tnJ_{i, t}^{(0)} = \tnH_{i, t - 1}$
($\tnH_{i, 0} = 0$ as defined later).
$\text{MSG}_\text{rc}^{(l)}$ and $\text{UPDATE}_\text{rc}^{(l)}$ are learnable
parametric functions of arbitrary forms, e.g., Multi-Layer Perceptions (MLPs).
Finally, the output for arbitrary node $i \in \stV$ is $
    \big[
        \text{GNN}_\text{rc}^{L}\big( \tnH_{:, t - 1}, \tnA_{:, :, t} \big)
    \big]_{i} := \tnJ_{i, t}^{(L)}
$.
\item
{\bf Description of $\text{Cell}$:}
For the sake of simplicity, we redefine $\tnH_{i, t}$ of \Cref{%
    eqn:timeandgraph%
} based on aformentioned symbols:
\begin{align}
\tnH_{i, t} = \text{Cell}\left( \tnI_{i, t}^{(L)}, \tnJ_{i, t}^{(L)} \right).
\label{eqn:timeandgraph_cell}
\end{align}
We initialize the corner case $\tnH_{i, 0} = 0$ for any node $i \in \stV$.
\end{enumerate}
\Cref{eqn:timeandgraph_in,eqn:timeandgraph_rc,eqn:timeandgraph_cell} give a
general description of existing architectures for \timeandgraph representation
functions.
Finally, $\tnH_{i, T}$ is the final temporal graph representation of node
$i \in \stV$, and is the target we want achieve from a constructed
\timethengraph representation.

Now, our target is to construct a \timethengraph representation function, and
we have to show the construction always give the same output $\tnH_{i, T}$ for
the same temporal graph input $
    \Big(
        \big[ \tnX_{i, t} \big]_{\forall i \in \stV, 1 \leq t \leq T},
        \big[ \tnA_{i, j, t} \big]_{\forall i, j \in \stV, 1 \leq t \leq T}
    \Big)
$.

First, we revisit the definition of \timethengraph representation in
\Cref{eqn:timethengraph}:
\begin{equation*}
\begin{aligned}
\tnH^\text{node}_{i} & = \text{RNN}^\text{node}\big( \tnX_{i, \leq T} \big),
\quad \forall i \in \stV,
\\
\tnH^\text{edge}_{i, j} & = \text{RNN}^\text{edge}\big(
    \tnA_{i, j, \leq T}
\big), \quad \forall (i, j) \in \stE,
\\
\tnZ & = \text{GNN}^{L}\big( \tnH^\text{node}, \tnH^\text{edge} \big).
\end{aligned}
\end{equation*}
\citet{siegelmann1995on} shows that with enough hidden neurons, a RNN can be
a most-expressive sequence model (universal Turing machine approximator).
Hence, since the representations for node and edge attribute sequences are
most-expressive, we can equivalently think of those RNN outputs simply as
``copy'' of the inputs, i.e., $\tnH^\text{node}_{i}$ perfectly represents $
    \tnX_{i, \leq T}
$ and $\tnH^\text{edge}_{i, j}$ perfectly represents $\tnA_{i, j, \leq T}$.
Hence, for the ease of simplicity, we can construct \timethengraph
representation just as a static GNN representation applied over the aggregated
node attribute and adjacency matrix inputs:
\begin{align*}
\tnZ' = \text{GNN}_\text{tar}^{TL}\big(
    \tnX_{:, \leq T}, \tnA_{:, :, \leq T}
\big),
\end{align*}
which we will describe construction details in the later content.

{
    \bf The constructed $\mathbf{\text{GNN}_\text{tar}^{TL}}$ representation
    details.
}
The basic idea is straightforward:
$\text{GNN}_\text{tar}^{TL}$ has $T \times L$ layers;
Between layers $
    l \in \{(\tau - 1) L + 1, \cdots, \tau L\}
$ of the construction, we will focus on emulating \timeandgraph
representation over node attributes $
    \tnX_{:, \tau}
$ and edges $\tnA_{:, :,  \tau}$ of snapshot $\tau = 1, \cdots, T$, which is
provided by the most-expressive sequence representations $\tnX_{:, \leq T}$ and
$\tnA_{:, :, \leq T}$ as our construction input.
In other words, constructed layer $l$ will work on data of snapshot $
    \lceil l / L \rceil
$ where $L$ is the number of GNN layers in the simulating \timeandgraph
representation.

In what follows, we extend some notations for the ease of construction
definition:
\begin{itemize}
\item
$\Vert$ denotes a vector concatenation operator;
\item
$l^{\%}$ denotes for a modified modulo
\begin{align*}
l^{\%} = \begin{cases}
    l \text{ mod } L, & l \text{ mod } L \neq 0,
    \\
    L, & l \text{ mod } L = 0;
\end{cases}
\end{align*}
\item
$
    \stN(i) = \{
        \forall j \in \stV
        |
        \exists \tau, 1 \leq \tau \leq T, \tnA_{i, j, \tau} > 0
    \}
$ denotes the set of full neighbors of $i \in \stV$ in the aggregated adjacency
matrix over all $T$ times;
\item
$\stN_{t}(i)$ is the neighbor set of node $i$ from $\tnA_{:, :, t}$ at snapshot
$t$.
\end{itemize}

Recall the \timeandgraph definitions in \Cref{%
    eqn:timeandgraph_in,eqn:timeandgraph_rc,eqn:timeandgraph_cell%
}, we subsitute their superscripts by $l^{\%}$ to fit our \timethengraph
construction (e.g., $
    \text{MSG}_\text{in}^{(l^{\%})}, \text{UPDATE}_\text{in}^{(l^{\%})},
    \text{MSG}_\text{rc}^{(l^{\%})}, \text{UPDATE}_\text{rc}^{(l^{\%})}
$).
We further use a tilde symbol to corresponding output variables (e.g., $
    \tnE, \tnI, \tnF, \tnJ
$) to reflect the simulating relationship in our construction (e.g., $
    \bld{\tnE}, \bld{\tnI}, \bld{\tnF}, \bld{\tnJ}
$).

Then, we design the output of each constructed layer $l \in \{1, \cdots, TL\}$
as a combination of three parts $
    \big[
        \tnX_{i, \leq T} \big\Vert \bld{\tnI}_{i}^{(l)} \big\Vert
        \bld{\tnJ}_{i}^{(l)}
    \big]
$ based on their usage
\begin{enumerate}
\item
$\tnX_{i, \leq T}$ is the attribute of node $i$ in aggregated temporal graph,
and it can also be regarded as the concatenation of all attributes of node $i$
of all time steps $1 \leq t \leq T$;
\item
$\bld{\tnI}^{(l)}_{i}, \bld{\tnJ}^{(l)}_{j}$ is used to achieve the same
representations as $
    \tnI^{(l^{\%})}_{i, \lceil l / L \rceil},
    \tnJ^{(l^{\%})}_{j, \lceil l / L \rceil}
$ in \Cref{eqn:timeandgraph_in,eqn:timeandgraph_rc}.
\end{enumerate}
For arbitrary node $i$ at constructon layer $l$, we need to constuct two
essential GNN components $\text{MSG}_\text{tar}^{(l)}$ and $
    \text{UPDATE}_\text{tar}^{(l)}
$ to properly pass these three parts as inputs and outputs.
First, $\text{MSG}_\text{tar}^{(l)}$ takes the concatenation of three parts of
arbitrary neighbor node $j \in \stN(i)$ as input:
Aggregated neighbor node attributes $\tnX_{j, \leq T}$;
And $\bld{\tnI}_{j}^{(l - 1)}, \bld{\tnJ}_{j}^{(l - 1)}$ that mimic $
    \tnI^{(l^{\%} - 1)}_{j, \lceil l / L \rceil},
    \tnJ^{(l^{\%} - 1)}_{j, \lceil l / L \rceil}
$ as in \Cref{eqn:timeandgraph_in,eqn:timeandgraph_rc}.
Then, $\text{UPDATE}_\text{tar}^{(l)}$ outputs $
    \big[ \bld{\tnE}_{i}^{(l)} \big\Vert \bld{\tnF}_{i}^{(l)} \big]
$ that mimic the representations of $
    \tnE^{(l^{\%})}_{i, \lceil l / L \rceil},
    \tnF^{(l^{\%})}_{i, \lceil l / L \rceil}
$ as in \Cref{eqn:timeandgraph_in,eqn:timeandgraph_rc}.

We also initialize the corner case
\begin{align}
\big[
    \tnX_{i, \leq T} \big\Vert \bld{\tnI}_{i}^{(0)} \big\Vert
    \bld{\tnJ}_{i}^{(0)}
\big] = \big[
    \tnX_{i, \leq T} \big\Vert \tnX_{i, 1} \big\Vert 0
\big], \label{eqn:construct_init}
\end{align}
for any node $i$.

{
    \bf In the following \Cref{eqn:construct_msg,eqn:construct_update} we give
    formal constructions of functions $\mathbf{\text{MSG}_\text{tar}^{(l)}}$
    and $\mathbf{\text{UPDATE}_\text{tar}^{(l)}}$ in $
        \mathbf{\text{GNN}_\text{tar}^{TL}}
    $ so that they can mimic all variables as discussed before:
}

\begin{equation}
\label{eqn:construct_msg}
\begin{aligned}
& \sum\limits_{j \in \stN(i)} \text{MSG}_\text{tar}^{(l)}\big(
    \big[
        \tnX_{i, \leq T} \big\Vert \bld{\tnI}_{i}^{(l - 1)} \big\Vert
        \bld{\tnJ}_{i}^{(l - 1)}
    \big], \big[
        \tnX_{j, \leq T} \big\Vert \bld{\tnI}_{j}^{(l - 1)} \big\Vert
        \bld{\tnJ}_{j}^{(l - 1)}
    \big], \tnA_{i, j, \leq T}
\big)
\\
& = \Bigg[
    \sum\limits_{
        j \in \stN_{\lceil \frac{l}{L} \rceil}(i)
    } \text{MSG}_\text{in}^{(l^{\%})}\big(
        \bld{\tnI}_{i}^{(l - 1)}, \bld{\tnI}_{j}^{(l - 1)},
        \tnA_{i, j, \lceil \frac{l}{L} \rceil}
    \big) \Bigg\Vert \sum\limits_{
        j \in \stN_{\lceil \frac{l}{L} \rceil}(i)
    } \text{MSG}_\text{rc}^{(l^{\%})}\big(
        \bld{\tnJ}_{i}^{(l - 1)}, \bld{\tnJ}_{j}^{(l - 1)},
        \tnA_{i, j, \lceil \frac{l}{L} \rceil}
    \big)
\Bigg]
\\
& = \big[ \bld{\tnE}_{i}^{(l)} \big\Vert \bld{\tnF}_{i}^{(l)} \big].
\end{aligned}
\end{equation}

\begin{equation}
\label{eqn:construct_update}
\begin{aligned}
& \text{UPDATE}_\text{tar}^{(l)}\big(
    \big[
        \tnX_{i, \leq T} \big\Vert \bld{\tnI}_{i}^{(l - 1)} \big\Vert
        \bld{\tnJ}_{i}^{(l - 1)}
    \big], \big[ \bld{\tnE}_{i}^{(l)} \big\Vert \bld{\tnF}_{i}^{(l)} \big]
\big)
\\
& = \begin{dcases}
    \Big[
        \tnX_{i, \leq T} \Big\Vert \text{UPDATE}_\text{in}^{(l^{\%})}\big(
            \bld{\tnI}_{i}^{(l - 1)}, \bld{\tnE}_{i}^{(l)}
        \big) \Big\Vert \text{UPDATE}_\text{rc}^{(l^{\%})}\big(
            \bld{\tnJ}_{i}^{(l - 1)}, \bld{\tnF}_{i}^{(l)}
        \big)
    \Big], & l \text{ mod } L \neq 0,
    \\
    \bigg[
        \tnX_{i, \leq T} \bigg\Vert \tnX_{i, \lceil \frac{l}{L} + 1 \rceil}
        \bigg\Vert \text{Cell}\Big(
            \text{UPDATE}_\text{in}^{(l^{\%})}\big(
                \bld{\tnI}_{i}^{(l - 1)}, \bld{\tnE}_{i}^{(l)}
            \big),
            \text{UPDATE}_\text{rc}^{(l^{\%})}\big(
                \bld{\tnJ}_{i}^{(l - 1)}, \bld{\tnF}_{i}^{(l)}
            \big)
        \Big)
    \bigg], & l \text{ mod } L = 0,
\end{dcases}
\\
& = \big[
    \tnX_{i, \leq T} \big\Vert \bld{\tnI}_{i}^{(l)} \big\Vert
    \bld{\tnJ}_{i}^{(l)}
\big].
\end{aligned}
\end{equation}

{
    \bf Finally, we are going to show the GNN construction defined by \Cref{%
        eqn:construct_msg,eqn:construct_update%
    } can give the same output $\mathbf{\tnH_{i, T}}$ as in \Cref{%
        eqn:timeandgraph_in%
    } for all nodes $\mathbf{i \in \stV}$.
}
Indeed, we will show by induction that $\bld{\tnJ}_{i}^{(tL)} = \tnH_{i, t}$
for arbitrary node $i$ at any time $t \in \{0, \cdots, T\}$.

First, as the initial point of induction, for any node $i \in V$ before the
first layer $l \leftarrow 1$, based on \Cref{eqn:construct_init}, we
initialize the input by
\begin{align*}
\big[
    \tnX_{i, \leq T} \big\Vert \bld{\tnI}_{i}^{(0)} \big\Vert
    \bld{\tnJ}_{i}^{(0)}
\big] = \big[ \tnX_{i, \leq T} \big\Vert \tnX_{i, 1} \big\Vert 0 \big].
\end{align*}
This implies that
\begin{align*}
\bld{\tnI}_{i}^{(0)} & = \tnX_{i, 1} = \tnI_{i, 1}^{(0)}, \quad \forall i \in
\stV,
\\
\bld{\tnJ}_{i}^{(0)} & = 0 = \tnJ_{i, 1}^{(0)} = \tnH_{i, 0}, \quad \forall i
\in \stV.
\end{align*}
Thus, $\bld{\tnJ}_{i}^{(tL)} = \tnH_{i, t}$ holds when $t \leftarrow 0$.

Now, suppose that
\begin{equation}
\label{eqn:induce_assume}
\begin{aligned}
\bld{\tnI}_{i}^{(l - 1)} & =
\tnI_{i, \lceil \frac{l}{L} \rceil}^{(l^{\%} - 1)}, \quad \forall i \in \stV,
\\
\bld{\tnJ}_{i}^{(l - 1)} & =
\tnJ_{i, \lceil \frac{l}{L} \rceil}^{(l^{\%} - 1)}, \quad \forall i \in \stV,
\end{aligned}
\end{equation}
for arbitrary $l \in \{1, \cdots, TL\}$.

Next, as the first step of induction, we show the message outputs of \Cref{%
    eqn:construct_msg%
} is the same as message outputs of \Cref{%
    eqn:timeandgraph_in,eqn:timeandgraph_rc%
} given the condition in \Cref{eqn:induce_assume}.
\begin{equation}
\label{eqn:induce_msg}
\begin{aligned}
\bld{\tnE}_{i}^{(l)} & =
\sum\limits_{j \in \stN_{\lceil \frac{l}{L} \rceil}(i)}
\text{MSG}_\text{in}^{(l^{\%})}\big(
    \bld{\tnI}_{i}^{(l - 1)}, \bld{\tnI}_{j}^{(l - 1)},
    \tnA_{i, j, \lceil \frac{l}{L} \rceil}
\big)
\\
& =  \sum\limits_{j \in \stN_{\lceil \frac{l}{L} \rceil}(i)}
\text{MSG}_\text{in}^{(l^{\%})}\big(
    \tnI_{i, \lceil \frac{l}{L} \rceil}^{(l^{\%} - 1)},
    \tnI_{j, \lceil \frac{l}{L} \rceil}^{(l^{\%} - 1)},
    \tnA_{i, j, \lceil \frac{l}{L} \rceil}
\big)
\\
& = \tnE_{i, \lceil \frac{l}{L} \rceil}^{(l^{\%})},
\\
\bld{\tnF}_{i}^{(l)} & =
\sum\limits_{j \in \stN_{\lceil \frac{l}{L} \rceil}(i)}
\text{MSG}_\text{rc}^{(l^{\%})}\big(
    \bld{\tnJ}_{i}^{(l - 1)}, \bld{\tnJ}_{j}^{(l - 1)},
    \tnA_{i, j, \lceil \frac{l}{L} \rceil}
\big)
\\
& = \sum\limits_{j \in \stN_{\lceil \frac{l}{L} \rceil}(i)}
\text{MSG}_\text{rc}^{(l^{\%})}\big(
    \tnJ_{i, \lceil \frac{l}{L} \rceil}^{(l^{\%} - 1)},
    \tnJ_{j, \lceil \frac{l}{L} \rceil}^{(l^{\%} - 1)},
    \tnA_{i, j, \lceil \frac{l}{L} \rceil}
\big)
\\
& = \tnF_{i, \lceil \frac{l}{L} \rceil}^{(l^{\%})}.
\end{aligned}
\end{equation}

Then, as the second step of induction, we show the update outputs of \Cref{%
    eqn:construct_update%
} is the same as update outputs of \Cref{%
    eqn:timeandgraph_in,eqn:timeandgraph_rc%
} given the condition in \Cref{eqn:induce_assume} and \Cref{eqn:induce_msg}.
Pay attention that there are two cases in \Cref{eqn:construct_update}, and we
will show those cases one by one.

When $l \text{ mod } L \neq 0$,
\begin{equation}
\label{eqn:induce_update1}
\begin{aligned}
\bld{\tnI}_{i}^{(l)} & = \text{UPDATE}_\text{in}^{(l^{\%})}\big(
    \bld{\tnI}_{i}^{(l - 1)}, \bld{\tnE}_{i}^{(l)}
\big)
\\
& = \text{UPDATE}_\text{in}^{(l^{\%})}\big(
    \tnI_{i, \lceil \frac{l}{L} \rceil}^{(l^{\%} - 1)},
    \tnE_{i, \lceil \frac{l}{L} \rceil}^{(l^{\%})}
\big)
\\
& = \tnI_{i, \lceil \frac{l}{L} \rceil}^{(l^{\%})}
= \tnI_{i, \lceil \frac{l + 1}{L} \rceil}^{(l^{\%})},
\\
\bld{\tnJ}_{i}^{(l)} & = \text{UPDATE}_\text{rc}^{(l^{\%})}\big(
    \bld{\tnJ}_{i}^{(l - 1)}, \bld{\tnF}_{i}^{(l)}
\big)
\\
& = \text{UPDATE}_\text{rc}^{(l^{\%})}\big(
    \tnJ_{i, \lceil \frac{l}{L} \rceil}^{(l^{\%} - 1)},
    \tnF_{i, \lceil \frac{l}{L} \rceil}^{(l^{\%})}
\big)
\\
& = \tnJ_{i, \lceil \frac{l}{L} \rceil}^{(l^{\%})} =
\tnJ_{i, \lceil \frac{l + 1}{L} \rceil}^{(l^{\%})}.
\end{aligned}
\end{equation}
We can see that \Cref{eqn:induce_update1} are indeed \Cref{eqn:induce_assume}
when $l \leftarrow l + 1$.
We provide a sketched illustration for this induction case in \Cref{%
    fig:layer_mid%
}.

When $l \text{ mod } L = 0$,
\begin{equation}
\label{eqn:induce_update2}
\begin{aligned}
\bld{\tnI}_{i}^{(l)} & = \tnX_{i, \lceil \frac{l}{L} + 1 \rceil} =
\tnI_{i, \lceil \frac{l + 1}{L} \rceil}^{(0)},
\\
\bld{\tnJ}_{i}^{(l)} & = \text{Cell}\big(
    \text{UPDATE}_\text{in}^{(l^{\%})}\big(
        \bld{\tnI}_{i}^{(l - 1)}, \bld{\tnE}_{i}^{(l)}
    \big),
    \text{UPDATE}_\text{rc}^{(l^{\%})}\big(
        \bld{\tnJ}_{i}^{(l - 1)}, \bld{\tnF}_{i}^{(l)}
    \big)
\big)
\\
& = \text{Cell}\big(
    \text{UPDATE}_\text{in}^{(l^{\%})}\big(
        \tnI_{i, \lceil \frac{l}{L} \rceil}^{(l^{\%} - 1)},
        \tnE_{i, \lceil \frac{l}{L} \rceil}^{(l^{\%})}
    \big),
    \text{UPDATE}_\text{rc}^{(l^{\%})}\big(
        \tnJ_{i, \lceil \frac{l}{L} \rceil}^{(l^{\%} - 1)},
        \tnF_{i, \lceil \frac{l}{L} \rceil}^{(l^{\%})}
    \big)
\big)
\\
& = \text{Cell}\big(
    \tnI_{i, \lceil \frac{l}{L} \rceil}^{(L)},
    \tnJ_{i, \lceil \frac{l}{L} \rceil}^{(L)}
\big)
\\
& = \tnH_{i, \lceil \frac{l}{L} \rceil} =
\tnJ_{i, \lceil \frac{l + 1}{L} \rceil}^{(0)}.
\end{aligned}
\end{equation}
We can see that \Cref{eqn:induce_update2} also corresponds to \Cref{%
    eqn:induce_assume%
} when $l \leftarrow l + 1$.
We provide a sketched illustration for this induction case in \Cref{%
    fig:layer_fin%
}.

We continue the induction until layer $l \leftarrow TL$, where we stop, and
\Cref{eqn:induce_update2} eventually yields
\begin{align*}
\bld{\tnJ}_{i}^{(TL)} = \tnH_{i, T}.
\end{align*}
This implies that we get the targeting \timeandgraph representation output $
    \tnH_{i, T}
$ from constructed (static) GNN output for an arbitrary node $i \in \stV$.

The above shows that {
    \bf any arbitrary \timeandgraph representation relying on 1-WL GNNs
    (\Cref{eqn:timeandgraph_in,eqn:timeandgraph_rc,eqn:timeandgraph_cell}) can
    be emulated by a \timethengraph representation (\Cref{%
        eqn:construct_msg,eqn:construct_update%
    }), which outputs the same representations for the same temporal graph
    inputs.
}
Thus, \timethenonewl is at least the same expressive as \timeandonewl based on
\Cref{lem:express_more_alter}.

%
{
    \bf Finally, we show a specific task (illustrated as \Cref{fig:dyncsl})
    where \timethenonewl is more expressive than \timeandonewl.
}

We now propose a synthetic task, whose temporal graph is illustrated in
\Cref{fig:dyncsl}.
The goal is to differentiate the topologies between two 2-step temporal graphs.
Each snapshot is a Circular Skip Link (CSL) graph (see \citet{%
    murphy2019relational%
}) with 7 unattributed nodes, denoted as $\gpC_{7, s}$, where $s$ represents
the smallest number of nodes on the outer circle between two neighbors which
are not connected by the outer circle.
For example, ``a'' and ``c'' are such two neighbors in $\gpC_{7, 1}$, and the
nodes on the outer circle between them is ``b'', which means $s = 1$.

In \Cref{fig:dyncsl}, two temporal graphs differ in their second time step $
    t_{2}
$.
From \citep{murphy2019relational}, if the CSL graphs are unattributed, any 1-WL
GNN will output the same representations for both $\gpC_{7, 1}$ and $
    \gpC_{7, 2}
$.
We use $\tnA^{(\text{top})}$ to represent the adjacency matrix of dynamics in
the top left of \Cref{fig:dyncsl}, and $\tnA^{(\text{btm})}$ for dynamics in
the bottom left of \Cref{fig:dyncsl}.
Note that $\tnX^{(\text{top})} = \tnX^{(\text{btm})} = 0$ since the temporal
graph is unattributed.

Hence, for a \timeandonewl representation,
\begin{align*}
& \text{GNN}_\text{in}^{L}\big(
    \tnX^{(\text{top})}_{:, 1}, \tnA^{(\text{top})}_{:, :, 1}
\big) = \text{GNN}_\text{in}^{L}\big(
    \tnX^{(\text{top})}_{:, 2}, \tnA^{(\text{top})}_{:, :, 2}
\big) = \text{GNN}_\text{in}^{L}\big(
    \tnX^{(\text{btm})}_{:, 1}, \tnA^{(\text{btm})}_{:, :, 1}
\big) = \text{GNN}_\text{in}^{L}\big(
    \tnX^{(\text{btm})}_{:, 2}, \tnA^{(\text{btm})}_{:, :, 2}
\big),
\\
& \text{GNN}_\text{rc}^{L}\big(
    \tnH^{(\text{top})}_{:, 0}, \tnA^{(\text{top})}_{:, :, 1}
\big) = \text{GNN}_\text{rc}^{L}\big( 0, \tnA^{(\text{top})}_{:, :, 1} \big) =
\text{GNN}_\text{rc}^{L}\big(
    \tnH^{(\text{btm})}_{:, 0}, \tnA^{(\text{btm})}_{:, :, 1}
\big) = \text{GNN}_\text{rc}^{L}\big( 0, \tnA^{(\text{btm})}_{:, :, 1} \big).
\end{align*}
Then, when we apply \Cref{eqn:timeandgraph_cell} at the first time step, we
will get exactly the same hidden representation $\tnH^{(\text{top})}_{:, 1}$
and $\tnH^{(\text{btm})}_{:, 1}$.
This also results in exactly the same hidden representation $
    \tnH^{(\text{top})}_{:, 2}
$ and $\tnH^{(\text{btm})}_{:, 2}$ following similar computations.
{
    \bf Thus, \timeandonewl will output the same final representation $
        \mathbf{\tnH^{(\text{top})}_{:, 2}}
    $ and $\mathbf{\tnH^{(\text{btm})}_{:, 2}}$ for two different temporal
    graphs in \Cref{fig:dyncsl}.
}

On the other hand, \timethenonewl will work directly on aggregated graph as
shown in the right side of \Cref{fig:dyncsl}.
Here, we can manually verify that a 1-WL GNN can distinguish these two
aggregated graphs, implying that a \timethenonewl representation can
distinguish these two temporal graphs that \timeandonewl cannot distinguish.

Finally, we conclude:
\begin{enumerate}
\item
The \timethenonewl is at least as expressive as the \timeandonewl;
\item
The \timethenonewl can distinguish temporal graphs not distinguishable by
\timeandonewl.
\end{enumerate}
Thus, \timethenonewl is strictly more expressive than \timeandonewl.
More precisely,
\begin{align*}
\text{{\Timeandonewl}} \exprlt_{\fmT^{|\stV|, T, p}} \text{{\Timethenonewl}},
\end{align*}
concluding our proof.

Besides, we show that \graphthentime is a strict subset of \timeandgraph by definition (\Cref{eqn:graphthentime,eqn:timeandgraph}), thus it is straightforward that \timeandonewl is strictly more expressive than \onewlthentime.

\end{proof}

\subsection{Temporal GNN Expressivity}
\label{subsec:proof_temp_graph_express}

\timegraph*

\begin{proof}

In the proof of \Cref{thm:temp_1wl_express}, we have noted that we can think
the RNNs of \timethengraph as capable of perfectly copying the sequence into a
representation, which ensures that the \timethengraph expressivity is
equivalent to the expressivity of a (static) GNN whose node and edge attributes
are their respective temporal sequences.

In \citet{morris2019weisfeiler}, it is shown that $
    \text{1-WL GNN} \exprlt_{\fmG_{|\stV|, p}} \text{$k$-WL GNN}
$.
And \citet{murphy2019relational,maron2019provably} both define most expressive
graph representations which we denote as GNN$^+$.

Using these known RNN and GNN expressivity results we have
\begin{align*}
\text{1-WL GNN} \exprlt_{\fmG^{|\stV|, p}} \text{$k$-WL GNN}
\exprlt_{\fmG^{|\stV|, p}} \text{GNN$^+$}
\end{align*}
respectively, which then yields
\begin{align*}
\text{\timethenonewl} \exprlt_{\fmT^{|\stV|, T, p}} \text{\timethenkwl}
\exprlt_{\fmT^{|\stV|, T, p}} \text{\timethenplus}
\end{align*}
and \timethenplus is the most expressive representation on $
    \fmT_{|\stV|, T, p}
$.

On the other hand, if we have the most expressive graph representation GNN$^+$,
then we can have an injection from any non-isomorphic snapshots to unique
representations.
Since RNN is also the most expressive for finite-length sequences, there also
exists an injection from finite-length sequence of snapshot representations to
a final \timeandplus representation.
Combining these two injections together, we can always get a representation
injection from temporal graph space $\fmT_{|\stV|, T, p}$, thus
\timeandplus is also the most expressive representation on $
    \fmT_{|\stV|, T, p}
$.
Hence, both \timethenplus and \timeandplus are most-expressive representations
of temporal attributed graphs with discrete time steps.

To summarize, we eventually have
\begin{align*}
\text{\timethenonewl} \exprlt_{\fmT_{|\stV|, T, p}} \text{\timethenkwl}
\exprlt_{\fmT_{|\stV|, T, p}} \text{\timethenplus}
\expreq_{\fmT_{|\stV|, T, p}} \text{\timeandplus}.
\end{align*}

\end{proof}

\subsection{TGAT and TGN Failiure on DynCSL}
\label{subsec:tgn_on_dyncsl}

Interestingly, TGAT and TGN, as a subset of \timethengraph representation (but
not \timeandgraph) baselines, they still do not work in the synthetic DynCSL
task just as \timeandgraph baselines (see \Cref{tab:classification}).
This poor performance is caused by the heterogeneous graph construction in TGAT
and TGN.
We will take TGN on \Cref{fig:dyncsl} for the case study, and TGAT will follow
similar explanation.

The first step of TGN is collecting all edges connecting to the same node over
all times as a sequence, and update temporal node attributes by the
representation of the collected sequence.
In two temporal graphs of \Cref{fig:dyncsl}, all nodes are unattributed and
their 1-hop topologies are always the same at every snapshot.
Take node ``a'' in the top temporal graph as an example.
\begin{enumerate}
\item
At the first step $t_{1}$, we will have its 1-hop topologies over all time as
\begin{align*}
\Big[
    \big( \tnX_{\text{b}, 1}, \tnA_{\text{a}, \text{b}, 1}, t_{1} \big),
    \big( \tnX_{\text{c}, 1}, \tnA_{\text{a}, \text{c}, 1}, t_{1} \big),
    \big( \tnX_{\text{f}, 1}, \tnA_{\text{a}, \text{f}, 1}, t_{1} \big),
    \big( \tnX_{\text{g}, 1}, \tnA_{\text{a}, \text{g}, 1}, t_{1} \big)
\Big].
\end{align*}
Since temporal graphs are unattributed, we can simply regard this as a sequence
of 4 same elements $t_{1}$.
Suppose the representation of this sequence is $h_{1}$.
\item
At the second step $t_{2}$, we will have its 1-hop topologies over all time as
\begin{equation*}
\begin{aligned}
\Big[
    & \big( \tnX_{\text{b}, 1}, \tnA_{\text{a}, \text{b}, 1}, t_{1} \big),
    \big( \tnX_{\text{c}, 1}, \tnA_{\text{a}, \text{c}, 1}, t_{1} \big),
    \big( \tnX_{\text{f}, 1}, \tnA_{\text{a}, \text{f}, 1}, t_{1} \big),
    \big( \tnX_{\text{g}, 1}, \tnA_{\text{a}, \text{g}, 1}, t_{1} \big),
    \\
    & \big( \tnX_{\text{b}, 2}, \tnA_{\text{a}, \text{b}, 2}, t_{2} \big),
    \big( \tnX_{\text{c}, 2}, \tnA_{\text{a}, \text{c}, 2}, t_{2} \big),
    \big( \tnX_{\text{f}, 2}, \tnA_{\text{a}, \text{f}, 2}, t_{2} \big),
    \big( \tnX_{\text{g}, 2}, \tnA_{\text{a}, \text{g}, 2}, t_{2} \big)
\Big].
\end{aligned}
\end{equation*}
Since temporal graphs are unattributed, we can simply regard this as a sequence
of 4 elements $t_{1}$ and 4 elements $t_{2}$.
Suppose the representation of this sequence is $h_{2}$.
\end{enumerate}
Since all nodes have the same topologies, thus the newly constructed node
attributes are only different between different times, and are the same for
nodes at the same snapshot.

Next, TGN will get GNN representation from heterogeneous graph constructed by
above node attributes $h_{1}$ and $h_{2}$.
We still take node ``a'' as the example.
The neighbor set of ``a'' in the heterogeneous graph will be
\begin{equation*}
\begin{aligned}
\Big\{\!\!\Big\{
    & \big( h_{1}, \tnA_{\text{a}, \text{b}, 1}, t_{1} \big),
    \big( h_{1}, \tnA_{\text{a}, \text{c}, 1}, t_{1} \big),
    \big( h_{1}, \tnA_{\text{a}, \text{f}, 1}, t_{1} \big),
    \big( h_{1}, \tnA_{\text{a}, \text{g}, 1}, t_{1} \big),
    \\
    & \big( h_{2}, \tnA_{\text{a}, \text{b}, 2}, t_{2} \big),
    \big( h_{2}, \tnA_{\text{a}, \text{c}, 2}, t_{2} \big),
    \big( h_{2}, \tnA_{\text{a}, \text{f}, 2}, t_{2} \big),
    \big( h_{2}, \tnA_{\text{a}, \text{g}, 2}, t_{2} \big)
\Big\}\!\!\Big\}.
\end{aligned}
\end{equation*}
Since temporal graphs are unattributed, this is indeed a multiset of 4 tuples $
    (h_{1}, t_{1})
$ and 4 tuples $(h_{2}, t_{2})$.
We can simply repeat this process on the bottom temporal graph, and can easily
find node ``a'' in the bottom temporal graph will eventually get the same
multiset.
When the neighbor multisets are the same, 1-WL GNN will always return the same
representations.
Thus, there is no way to differeniate these two temporal graphs by TGN.

This case study explains the poor performance of TGAT and TGN on DynCSL
dataset.

\begin{figure}[tb]
\vskip 0.2in
\begin{center}
\centerline{\includegraphics[width=\textwidth]{fig_link_pred}}
\caption{
    {\bf Equivariant expressivity insufficiency instance for temporal graph.}
    A simple example of two disconnected components on a temporal knowledge
    graph without attributes, boreal forest (orange) and antarctic fauna (blue)
    over two-time dynamics in a food chain system.
    The two disconnected components have the same dynamics because of similar
    prey and predator environments in the isolated ecosystems.
    Extending \citet{srinivasan2020on} to temporal graphs, we can show that
    equivariant node representations of both \timethengraph and \timeandgraph
    will output the same temporal representations for solid gray nodes
    ``Coyote'' and ``Seal'' for the same temporal structure at $t_{2}$.
    Hence, temporal link prediction based on these node representations will
    predict the same predator-prey edges, e.g., with solid black ``Lynx'',
    where predicted dash edge between ``Seal'' and ``Lynx'' (which implies lynx
    also eat seal) is false positive.
}
\label{fig:link_pred}
\end{center}
\vskip -0.2in
\end{figure}

\subsection{Expressivity Insufficency For Link Prediction}
\label{subsec:proof_link_pred}

In this subsection, we give more details about the expressivity insufficiency
introduced in \Cref{sec:experiments}.
We focus the expressivity study mostly on \timethengraph, since we have shown
that \timethengraph is at least of the same expressvity as \timeandgraph as
\Cref{thm:temp_1wl_express} and \Cref{thm:temp_graph_express} with any kinds of
GNNs.
We start with the statement that an equivariant graph representation is a
structural node representation, since a permutation acts on the nodes (refer
\citet{you2019position,srinivasan2020on} for a description of the difference
between structural and positional node representations).

\Cref{fig:link_pred} shows a temporal graph with two disconnected components
(orange and blue) in a food web with two isolated ecosystems.
At each time step, the two disconnected components have the same topology.
Then, the sequences of snapshots of the aggregated temporal graphs will be the
same in these two disconnected components.
This means that the \timethengraph aggregated forms of the two disconnected
components (forest and sea) are exactly the same at time $t_{2}$.
We can now invoke \citep{srinivasan2020on} for static attributed graphs to
declare that the ``Coyote'' and ``Seal'' will receive the same \timethengraph
most expressive and equivariant node representation at time $t_{2}$.
By \Cref{thm:temp_graph_express}, the most expressive \timethengraph
representation is as expressive as the most expressive \timeandgraph
representation, thus ``Coyote'' and ``Seal'' will also have the same
\timeandgraph equivariant node representation.

Since ``Coyote'' and ``Seal'' have the same temporal node representations, the
method will incorrectly predict predatory links between two isolated ecosystem.
For instance, if we want to predict their relationship with ``Lynx'', node
representation based methods will predict both/none of ``Coyote'' and ``Seal''
have link to ``Lynx'' at second step.
The true answer is that ``Coyote'' has link to ``Lynx'' (directly shown in
\Cref{fig:link_pred} observation), but ``Seal'' should have no link to ``Lynx''
(in two isolated ecosystems).
So, the prediction based on such equivariant node representations will always
give wrong prediction at least for one animal kind.
Thus, equvariant \timeandgraph and \timethengraph temporal node representations
are insufficiently expressive to predict temporal links.

\section{Experiment Configurations}
\label{sec:configs}

\begin{table}[t]
\caption{
    {\bf Model Designs.}
    We introduce model architecture details in this table.
    Representation corresponds to the temporal graph representation families
    introduced in \Cref{sec:prelim}.
    Node RNN corresponds to the sequence representation applies on node-level
    data.
    Edge RNN corresponds to the sequence representation applies on edge-level
    data.
    GNN corresponds to the graph representation applies on graph data (either
    snapshots or aggregation).
    Other corresponds to implementation details which can not be simply
    clarified in this table, and is explained in \Cref{subsec:models}.
}
\label{tab:design}
\vskip 0.15in
\begin{center}
\resizebox{\columnwidth}{!}{
\begin{tabular}{lcllll}
\hline
\multicolumn{1}{c}{Model} & Representation & \multicolumn{1}{c}{Node RNN} &
\multicolumn{1}{c}{Edge RNN} & \multicolumn{1}{c}{GNN} &
\multicolumn{1}{c}{Other} \\
\hline
EvolveGCN-O & \multirow{4}{*}{\graphthentime} &
LSTM~\citep{hochreiter1997long} & \emph{N/A} & GCN~\citep{kipf2016semi} &
Recurrent$_\text{O}$ Weight \\
EvolveGCN-H & & GRU~\citep{chung2014empirical} & \emph{N/A} & GCN~\citep{%
    kipf2016semi%
} & Recurrent$_\text{H}$ Weight \\
GCN-GRU & & GRU~\citep{chung2014empirical} & \emph{N/A} & GCN~\citep{%
    kipf2016semi%
} & \\
DySAT & & GAT~\citep{velickovic2018graph} & \emph{N/A} & Attn~\citep{%
    vaswani2017attention%
} & \\
\hline
GCRN-M2 & \multirow{2}{*}{\timeandgraph} & LSTM~\citep{hochreiter1997long} &
\emph{N/A} & Spectral~\citep{defferrard2016convolutional} & \\
DCRNN & & GRU~\citep{chung2014empirical} & \emph{N/A} & Spectral~\citep{%
    defferrard2016convolutional%
} & Concatenate \\
\hline
TGAT & \multirow{3}{*}{\timethengraph} & \emph{N/A} & Heterogeneous &
GAT~\citep{velickovic2018graph} & Relative \\
TGN & & Last & Heterogeneous & GAT~\citep{velickovic2018graph} &
Incremental \\
GRU-GCN & & GRU~\citep{chung2014empirical} & GRU~\citep{chung2014empirical} &
GCN~\citep{kipf2016semi} & Skip \\
\hline
\end{tabular}
}
\end{center}
\vskip -0.1in
\end{table}

\subsection{Our Model and Baselines}
\label{subsec:models}

We summarize all architectures of our model and baselines in \Cref{tab:design}.
It covers most of the details of architecture designs.
However, there are still some implementation details which can not be simply
explained in \Cref{tab:design}.
We will briefly introduce those implementation details in later paragraphs.

{\bf Recurrent Weight.}
In both EvolveGCN-O and EvolveGCN-H, the hidden variable of RNNs is the
weight paremeters of their GNNs rather than the representations of history
snapshots.
Thus, RNN will model the evolution of GNN parameters, rather than snapshot
attributes.
At each snapshot, GNN is initialized by parameters given by RNNs, and only
receives current node and edge attributes as inputs to get final
representations.
The difference between EvolveGCN-O and EvolveGCN-H is that RNN of EvolveGCN-O
recurrently outputs GNN parameters without receiving any inputs, while RNN of
EvolveGCN-H takes node attributes $\tnX_{:, t}$ as inputs and models parameter
evolution with respect to node attributes.
We use different subscripts in \Cref{tab:design} to distinguish the RNN
difference between two models.

{\bf Concatenate.}
In DCRNN, instead of apply two GNNs independently on $
    \big( \tnX_{:, t}, \tnA_{:, :, t} \big)
$ and $\big( \tnH_{:, t - 1}, \tnA_{:, :, t} \big)$ as \Cref{eqn:timeandgraph},
it first concatenates $\tnX_{:, t}$ and $\tnH_{:, t - 1}$ of
the same nodes together as new node attributes, then get GNN representations on
the snapshot with concatentated node attributes and edge attributes $
    \tnA_{:, :, t}
$ as RNN outputs $\tnH_{:, t}$.

{\bf Heterogeneous.}
In both TGAT and TGN, it will collect all the edges connected to the same nodes
together and construct a heterogeneous graph for GNNs to embed.
For example, suppose for node $1$ we have following edges and neighbors over
time:
Edge $\tnA_{1, 2, 1}$ from node $2$ at time $t_{1}$, $\tnA_{1, 3, 1}$ from node
$3$ at time $t_{1}$, and edge $\tnA_{1, 3, 2}$ from node $3$ at time $t_{2}$.
Then, in the static heterogeneous graph, we will have following three edges:
Edge from $2$ to $1$ with attribute $(\tnA_{1, 2, 1}, t_{1})$, edge from $3$ to
$1$ with attribute $(\tnA_{1, 3, 1}, t_{1})$ and another edge from $3$ to $1$
with attribute $(\tnA_{1, 3, 2}, t_{2})$.
Timestamp data is added as augmented edgee attributes to distinguish edges from
different snapshots.

{\bf Relative.}
In TGAT, timestamp data of an arbitrary snapshot $t$ is achieved by the
relative time gap between current snapshot $t$ and the last snapshot $T$ in the
temporal graph.
If the raw timestamp is discrete, then the relative time gap should be $T - t$.

{\bf Last.}
In TGN, besides collecting edge attributes for heterogeneous graph, it also
collects new node attributes for heterogeneous graph.
The node attributes in heterogeneous graph of TGN is given by the latest edge
connecting to nodes.
For instance, suppose we are focusing on the same node $1$ as the example in
``Heterogeneous'' paragraph, since the latest edge comes from node $3$ at time
$t_{2}$, then the new node attributes will be $
    (\tnX_{1, 2}, \tnX_{3, 2}, \tnA_{1, 3, 2}, t_{2})
$ where $\tnX_{1, 2}, \tnX_{3, 2}$ are the node attributes of $1$ and $3$ at
time $t_{2}$.

{\bf Incremental.}
In TGN, timestamp data of an arbitrary snapshot $t$ is achieved by the
incremental time gap between the current snapshot $t$ and previous snapshot
$t - 1$ in the temporal graph.
If the raw timestamp is discrete, then the incremental time gap will always be
1 except for the first snapshot which is 0.

{\bf Skip.}
As shown in the expressivity proof \Cref{subsec:proof_temp_1wl_express}, we
need to connect the raw node inputs of GNN to the outputs to acheive the
maximum expressivity.
As a counterpart of this, in our GRU-GCN proposal, we concatenate the node
representation given by RNN to the output of GCN as a skip link.

In the experiments, the temporal graph representation space is fixed to be $
    \fmR^{16}
$.
For both classification and regression tasks, we use a MLP with 1 hidden layer
after the temporal graph representations to get the final predictions.
We use softplus as our activation function for all models except for DynCSL, we
use tanh which converges faster.

\begin{table*}[t]
\caption{
    {\bf Complexity Table.}
    The time complexity of all models evaluated in our work (in big O
    notation).
    $T$ is number of time steps, $V$ is number of nodes in temporal graph,
    $\sum_{t} E_{t}$ is total number of edges of all snapshots, $E_\text{agg}$
    is the number of edges in aggregated temporal graph and $d$ is
    representation dimension.
}
\label{tab:complexity}
\begin{center}
\begin{tabular}{lr}
\hline
\multicolumn{1}{c}{Model} & \multicolumn{1}{c}{Complexity} \\
\hline
EvolveGCN-O & $\mathcal{O} \big( T d^{2} + T V d^{2} + \sum_{t} E_{t} d \big)$ \\
EvolveGCN-H & $\mathcal{O} \big( T d^{2} + T V d^{2} + \sum_{t} E_{t} d \big)$ \\
GCN-GRU     & $\mathcal{O} \big( T d^{2} + T V d^{2} + \sum_{t} E_{t} d \big)$ \\
DySAT       & $\mathcal{O} \big( T d^{2} + T V d + \sum_{t} E_{t} d^{2} \big)$ \\
\hline
GCRN-M2     & $\mathcal{O} \big( T d^{2} + T V d^{2} + \sum_{t} E_{t} d \big)$ \\
DCRNN       & $\mathcal{O} \big( T d^{2} + T V d^{2} + \sum_{t} E_{t} d \big)$ \\
\hline
TGAT        & $\mathcal{O} \big( T V d + \sum_{t} E_{t} d^{2} \big)$ \\
TGN         & $\mathcal{O} \big( \sum_{t} E_{t} d^{2} + T V d^{2} + E_\text{agg} d \big)$ \\
GRU-GCN     & $\mathcal{O} \big( T E_\text{agg} d^{2} + T V d^{2} + E_\text{agg} d \big)$ \\
\hline
\end{tabular}
\end{center}
\vskip -0.1in
\end{table*}

\begin{table*}[t]
\caption{
    {\bf Performance without neighbor information.}
    We compare the performance of our proposal with or without using neighbor
    informatation, denoted as GRU-GCN and GRU respectively.
    It is obviously worse when we remove neighbor information from our
    proposal.
    This shows that temporal regression tasks can not be simply predicted with
    only node attribute dynamics.
}
\label{tab:pureseq}
\vskip 0.15in
\begin{center}
\resizebox{\textwidth}{!}{
\begin{tabular}{lrrrrrrrr}
\hline
\multicolumn{1}{c}{\multirow{2}{*}{Model}} &
\multicolumn{2}{c}{PeMS04} & \multicolumn{2}{c}{PeMS08} &
\multicolumn{2}{c}{Spain-COVID} & \multicolumn{2}{c}{England-COVID} \\
& \multicolumn{1}{c}{Transductive} & \multicolumn{1}{c}{Inductive} &
\multicolumn{1}{c}{Transductive} & \multicolumn{1}{c}{Inductive} &
\multicolumn{1}{c}{Transductive} & \multicolumn{1}{c}{Inductive} &
\multicolumn{1}{c}{Transductive} & \multicolumn{1}{c}{Inductive} \\
\hline
GRU-GCN &
$1.61 {\scriptstyle \pm 0.35}$\% &
$1.13 {\scriptstyle \pm 0.05}$\% &
$1.27 {\scriptstyle \pm 0.21}$\% &
$0.89 {\scriptstyle \pm 0.07}$\% &
$1.66 {\scriptstyle \pm 0.63}$\% &
$0.65 {\scriptstyle \pm 0.16}$\% &
$3.41 {\scriptstyle \pm 0.28}$\% &
$2.87 {\scriptstyle \pm 0.19}$\% \\
GRU &
$4.36 {\scriptstyle \pm 2.29}$\% & $2.09 {\scriptstyle \pm 0.61}$\% &
$3.22 {\scriptstyle \pm 1.31}$\% & $1.69 {\scriptstyle \pm 0.61}$\% &
$2.79 {\scriptstyle \pm 0.11}$\% & $2.09 {\scriptstyle \pm 0.06}$\% &
$5.89 {\scriptstyle \pm 0.04}$\% & $5.68 {\scriptstyle \pm 0.07}$\% \\
\hline
\end{tabular}
}
\end{center}
\vskip -0.1in
\end{table*}

\subsection{Complexity Analysis}
\label{subsec:complexity}

\Cref{tab:complexity} analyzes the complexity of all models in big O notations.
For an arbitrary temporal graph, we denote number of snapshots as $T$, number
of nodes as $V$, total number of edges as $\sum_{t} E_{t}$, number of edges in
aggregated temporal graph as $E_\text{agg}$ and $d$ is representation
dimension.
Furthermore, since all node and edge attributes are quite small (see \Cref{%
    tab:meta%
}), we can assume all attributes have dimension $\mathcal{O}(1)$.
We specially differentiate total number of snapshot edges $\sum_{t} E_{t}$ and
number of aggregated edges $E_\text{agg}$ since $E_\text{agg}$ may vary a lot
from $E_\text{agg} \ll \sum_{t} E_{t}$ to $
    E_\text{agg} \approx \sum_{t} E_{t}
$ (see \Cref{fig:dyncsl} as an example).
Our analysis is based on the following complexity assumptions for sequence and
graph representations.

\begin{itemize}
\item
Both GRU and LSTM have complexity $\mathcal{O}\big( T d^{2} \big)$ if the input
sequence has length $T$.
\item
Self attention mechanism has complexity $
    \mathcal{O}\big( T^{2} d + T d^{2} \big)
$ if the input sequence has length $T$, and since $T$ is quite small in all
experiments (see \Cref{tab:meta}), it is further simpified as $
    \mathcal{O}\big( T d^{2} \big)
$.
\item
For graph representations, we assume that each snapshot is sparse, in other
words, we can assume the node degree in any snapshot being $\mathcal{O}(1)$.
\item
Both GCN and SpectralGCN have complexity $\mathcal{O}\big( V d^{2} + E d \big)$
if the input graph has $V$ nodes and $E$ edges.
\item
GAT has complexity $\mathcal{O}\big( V d + E d^{2} \big)$ if the input graph
has $V$ nodes and $E$ edges.
\end{itemize}

\subsection{Learning Configurations}
\label{subsec:hypers}

{\bf Dataset split.}
In all experiments, datasets are split into 70\% for training, 10\% for
validation, and 20\% for test.
For transductive tasks, the split is based on nodes, and we ensure that the
degree in aggregated temporal graph (sum over all temporal graphs) has nearly
the same distribution among training, validation and test.
For inductive tasks, the split is simply based on chronological order that the
first 70\% temporal graphs in the dataset become training, next 10\% become
validation, and remaining 20\% become test.
We will normalize each attributes so that normalized attributes in training is
always between 0 and 1.

{\bf Evaluation Metrics.}
For temporal node and graph classification tasks (DynCSL and Brain10), we use
ROCAUC score~\citep{hand2001simple} as the evaluation metric.
For temporal node regression tasks (PeMS and COVID), we use mean average
percentage error (MAPE) as the evaluation metric, and it is formally defined as
\begin{align}
\frac{1}{N} \sum\limits_{n = 1}^{N} \frac{1}{V} \sum\limits_{v = 1}^{V}
\frac{1}{D} \sum\limits_{d = 1}^{D} \frac{
    \big| \hat{y}_{v, p}^{(n)} - y_{v, p}^{(n)} \big|
}{\big| y_{v, p}^{(n)} \big| + 1}. \label{eqn:mape}
\end{align}
where $\hat{y}_{v, p}^{(n)}$ is the model prediction for $p$-th attribute of
node $v$ in $n$-th temporal graph, and $y_{v, p}^{(n)}$ is the ground
truth for $p$-th attribute of node $v$ in $n$-th temporal graph.
$N$ is number of test temporal graphs, $V$ is number of test nodes, and $D$
is the dimension of node attributes.
Pay attention that we add 1 to denominator since $y_{v, p}^{(n)}$ may be 0 in
our datasets.
For PeMS, $D = 3$.
For COVID, $D = 1$.
We also transform the result into percentage notation in \Cref{tab:regression}.

{\bf Hyperparameter.}
In our experiments, we do a grid hyperparameter search for learning rates from
$0.1$, $0.01$ and $0.001$.
For each learning rate configuration, we run 10 times and collect corresponding
mean performance, and select the best configuration according to the mean
performance on validation set.
Then, in \Cref{tab:classification} and \Cref{tab:regression}, we report
evaluation results of selected configratuions on test data .

For the simplest task DynCSL, we train all methods by 30 epochs.
For the largest task Brain10, we train by 200 epochs to ensure convergence of
all methods.
On PeMS and COVID datasets, we train by 100 epochs.

\subsection{Real-world Datasets}
\label{subsec:datasets}

{\bf Brain10.}
Brain10 is based on a fMRI brain scans in a short period.
Nodes are voxels in the scan, and temporal edges are constructed by voxels
activation over time.
The goal is to predict the functionality category of each voxel given the full
dynamics.

{\bf PeMS.}
PeMS is a traffic forecasting task.
Each data point in PeMS is a temporal graph of 13 snapshots.
Each temporal node corresponds to a road sensor, and collects average traffic
statistics (flow, occupancy and speed) every 5 minutes.
Edges are defined by the geographic distance between two sensors.
The goal is to predict traffic statistics (flow, occupancy and speed) of the
last snapshot given the first 12 snapshots (past 1 hour) for a given data
point.
The hour and weekday information of the first 12 snapshots are also provided as
augmented inputs for the prediction.
The difference between PeMS04 and PeMS08 is that they are collected from
different districts of California at different months.
Our PeMS is different from \citet{guo2019attention} where only one attribute
(flow) of multiple future snapshots is predicted, while we only care about the
prediction of one future snapshot, but for all 3 attributes (flow, occupancy
and speed).
Furthermore, we additionally add timestamp data as augmented node inputs.

{\bf COVID.}
COVID is a COVID infection rate forecasting task.
Each data point in COVID is a temporal graph of 8 snapshots.
Each temporal node corresponds to a city, and collects new infection population
every day.
Edges are constructed by transportation populations and types (if possible)
between cities every day.
The goal is to predict infection of the last snapshot given the first 7
snapshots (past week).
The difference between Spain-COVID and England-COVID is that they are collected
in different countries.
Our SpainCOVID is different from \citet{panagopoulos2020transfer} where
attributes of multiple future snapshots are predicted, while we only care about
the prediction of one future snapshot.

\subsection{Computation Efficency}
\label{subsec:rescost}

\begin{figure}[tb]
\vskip 0.2in
\begin{center}
\centerline{\includegraphics[width=\columnwidth]{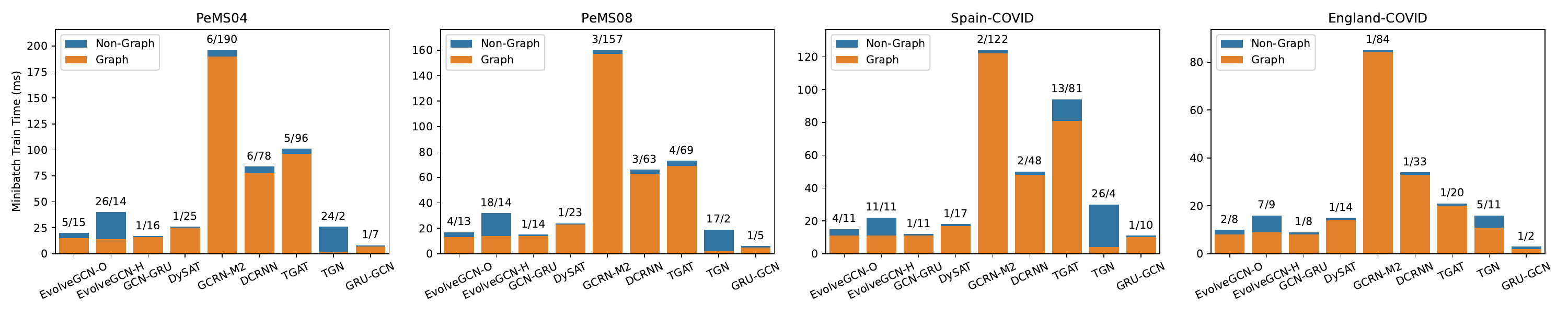}}
\caption{
    {\bf Runtime Details.}
    The orange bar corresponds to training time per minibatch spending on GNNs.
    The blue bar corresponds to training time per minibatch except for GNNs.
    The values ahead of each bar correspond to time cost of each color.
    The first one corresponds to blue bar (non-graph), and the other one
    corresponds to the orange bar (graph).
    We can see that GNNs occupy the majority of training time cost.
}
\label{fig:rescost}
\end{center}
\vskip -0.2in
\end{figure}

\begin{table*}[t]
\caption{
    {\bf CPU runtime on Brain10.}
    We show the CPU runtime on graph and non-graph operations on Brain10.
    We also provide number of edges used in graph convolutions in ``Conv.
    Edges'' column.
    We can see that on Brain10, we have a large amount of edges, and
    aggregation of our proposal does not significantly reduce the number of
    edges.
    Thus the GNN computation cost even increases since the edge attributes are
    more complex than \timeandgraph baselines.
    Furthermore, since we have a large amount of edges, the time cost for RNN
    on edge attributes of our proposal is also extremely high.
}
\label{tab:cpucost}
\vskip 0.15in
\begin{center}
\begin{tabular}{clrrr}
\hline
Representation & \multicolumn{1}{c}{Model} & Graph & Non-Graph & Conv. Edges \\
\hline
\graphthentime & EvolveGCN-O &  557 ms &    21 ms & 1955488 \\
\graphthentime & EvolveGCN-H &  533 ms &    57 ms & 1955488 \\
\graphthentime &     GCN-GRU &   51 ms &    26 ms & 1955488 \\
\graphthentime &       DySAT &  150 ms &    43 ms & 1955488 \\
\hline
\timeandgraph &     GCRN-M2 & 5940 ms &     7 ms & 1955488 \\
\timeandgraph &       DCRNN & 3542 ms &    17 ms & 1955488 \\
\hline
\timethengraph &        TGAT & 7282 ms &    89 ms & 1955488 \\
\timethengraph &         TGN & 2376 ms &   409 ms & 1955488 \\
\timethengraph &     GRU-GCN &  975 ms & 14638 ms & 1761414 \\
\hline
\end{tabular}
\end{center}
\vskip -0.1in
\end{table*}

We show the training runtime proportions of GNNs and non-GNNs for each model in
\Cref{fig:rescost}.

We can see the GCRN-M2 and DCRNN as the slowest two methods, spend nearly all
of their time on GNNs.
This is because they have multiple GNNs as \Cref{eqn:timeandgraph} while all
the other methods only have a single GNN.
For example, for GCRN-M2 whose sequence representation is LSTM, it will have 8
different GNNs:
It has two GNNs for input gate, forget gate, output gate and cell unit
respectively.

In the constrast, \graphthentime models (EvolveGCNs, GCN-GRU and DySAT) as a
subset of \timeandgraph, is faster.
This is because they only have GNN on input snapshots, and do not have GNN
depending on hidden representation from previous snapshots (see \Cref{%
    eqn:graphthentime%
}).
Thus, \graphthentime can compute GNN for different snapshots in the same time.
Besides, it does not require multiple GNNs as in \timeandgraph models.
These two advantages result in a great improvement on efficiency comparing with
\timeandgraph baselines (GCRN-M2, DCRNN).

Our proposal GRU-GCN spends very small amount of runtime on GNNs.
This is because GRU-GCN is applied on the aggregated temporal graph which
has far less amount of edges comparing to snapshotted temporal graphs in those
datasets.
In the meanwhile, our proposal GRU-GCN also spends small amount of runtime on
non-graph operations, e.g., RNNs, which eventually results in its great
efficiency.
Pay attention that RNN on edge attributes has small cost only because we are
using simple GRU representation and the number of edges are small on studying
datasets.
However, this is not true when we have large amount of edges. e.g., when the
temporal graph is dense.

Indeed, \Cref{tab:cpucost} finds that GRU-GCN is slowest on Brain10 where the
number of edges is very large and aggregation does not effectively reduce the
number of edges (edges are almost the same as snapshotted form).
Although \Cref{tab:cpucost} is runtimes on CPU rather than GPU, it still
reveals that the efficiency of \timethengraph is dependent on temporal graph
topologies, and it is not always true that \timethengraph will be more
efficient than \timeandgraph.

\end{document}